\documentclass[10pt,twocolumn,letterpaper]{article}
\usepackage{cvpr}              
\usepackage[dvipsnames]{xcolor}
\usepackage{booktabs}
\usepackage{graphicx}
\usepackage{multirow}
\usepackage{tabularx}
\usepackage{pifont}
\usepackage{float}

\definecolor{cvprblue}{rgb}{0.21,0.49,0.74}
\usepackage[pagebackref,breaklinks,colorlinks,citecolor=cvprblue]{hyperref}

\def\paperID{8531} 
\def\confName{CVPR}
\def\confYear{2024}

\title{Dynamic Inertial Poser (DynaIP): Part-Based Motion Dynamics Learning for Enhanced Human Pose Estimation with Sparse Inertial Sensors}

\author{Yu Zhang$^1$\thanks{Equal contribution}
\quad
Songpengcheng Xia $^{1*}$
\quad
Lei Chu $^{2}$ \thanks{Corresponding authors \\ This work was supported by the National Nature Science Foundation of China (NSFC) under Grant 62273229.} 
\quad
Jiarui Yang$^1$
\quad
Qi Wu$^1$
\quad
Ling Pei$^{1\dagger}$ 
\\
$^1$Shanghai Key Laboratory of Navigation and Location Based Services, Shanghai Jiao Tong University
\\
$^2$Wireless Devices and Systems Group (WiDeS), University of Southern California
}
\begin{document}

\maketitle


\begin{abstract}

This paper introduces a novel human pose estimation approach using sparse inertial sensors, addressing the shortcomings of previous methods reliant on synthetic data. It leverages a diverse array of real inertial motion capture data from different skeleton formats to improve motion diversity and model generalization. This method features two innovative components: a pseudo-velocity regression model for dynamic motion capture with inertial sensors, and a part-based model dividing the body and sensor data into three regions, each focusing on their unique characteristics. The approach demonstrates superior performance over state-of-the-art models across five public datasets, notably reducing pose error by 19\% on the DIP-IMU dataset, thus representing a significant improvement in inertial sensor-based human pose estimation. Our codes are available at {\url{https://github.com/dx118/dynaip}}



\end{abstract}    
\section{Introduction}
\label{sec:intro}
Human Pose Estimation (HPE) has emerged as a critical field of study, attracting considerable interest for its applications in various domains~\cite{jiang2022avatarposer,zheng2023deep}, like aiding in sports training and analysis, and enriching interactions in Virtual and Augmented Reality (VR/AR) environments. This growing relevance underscores the importance of advancing HPE technologies to meet the diverse needs of these applications. 

\begin{figure}[h]
	\includegraphics[width=8.3cm]{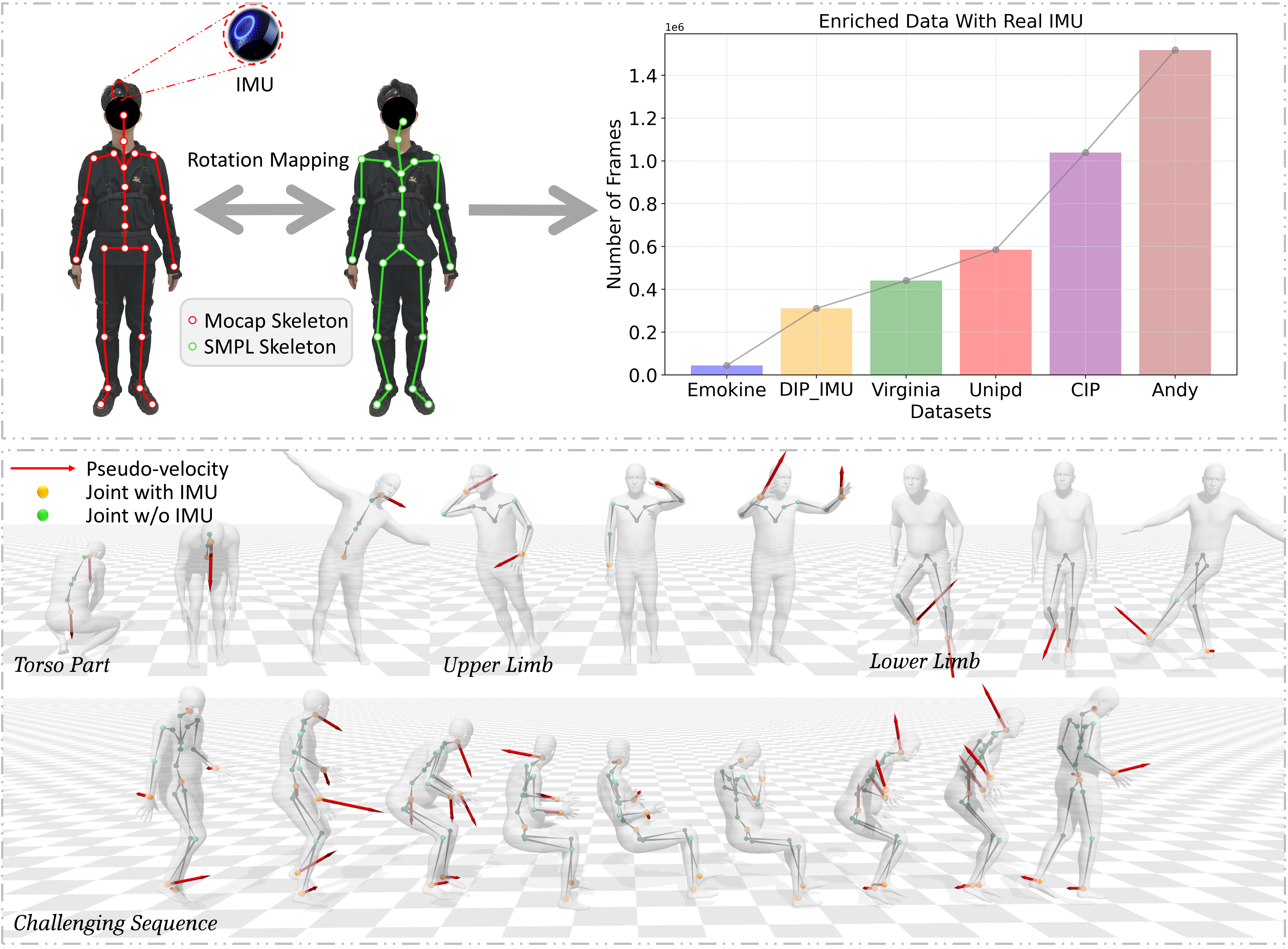}
	\caption{{Our innovative data-driven approach for robust full-body pose estimation using six IMUs: unifying inertial mocap datasets across skeleton formats and enhancing challenging motion capture with local body region modeling and pseudo-velocity estimation.}}
	\label{introduction}
	\vspace{-5mm}
\end{figure}

Our paper examines HPE, a field marked by varied sensing modalities and methodologies~\cite{an2022mri, guzov2021human, dai2023sloper4d}, divided into three categories: 1) Vision-based HPE~\cite{tripathi20233d, shen2023learning}, using single or multi-view images, known for its significant advancements; 2) Wireless-based HPE~\cite{zhao2018through, chen2022mmbody}, which addresses some vision-based challenges but is limited by environmental factors; 3) Wearable-based HPE~\cite{huang2018deep, kaufmann2021pose, yi2022physical}, our focus, which utilizes Inertial Measurement Units (IMUs) for full-body estimation. Vision-based HPE, despite its success, struggles with occlusion, privacy, and perspective issues. Wireless-based methods~\cite{zhao2018through, chen2022mmbody} mitigate some of these issues but face performance inconsistencies. Our study centers on wearable-based HPE, harnessing wearables with sensors for joint orientation and acceleration tracking. This IMU-based method excels by being occlusion-resistant, environmentally stable, and privacy-conscious, thus addressing the drawbacks of other HPE approaches.

Recent studies~\cite{huang2018deep, yi2021transpose, mollyn2023imuposer}  have concentrated on minimizing the number of IMU sensors for human motion reconstruction, balancing user convenience with less intrusiveness. While this has shown potential, the limited sensor placement leads to uncertainties in estimating unseen joint rotations. The integration of complex network structures ~\cite{jiang2022transformer} and physical constraints ~\cite{yi2022physical} has been proven effective in reducing ambiguities. Despite these developments, challenges persist, especially in complex motions. Our research aims to enhance the robustness of sparse sensor-based full-body pose reconstruction by identifying three key areas for potential improvements.

\textbf{Multi-modal Sensor Information Utilization}: 
The primary challenge in IMU-based HPE lies in optimally utilizing multi-modal sensor data, especially acceleration, to diminish motion ambiguity. Sole reliance on IMU orientation data for full-body pose estimation can result in ambiguities, particularly in movements like hand or leg raises. Acceleration data, with its dynamic motion features, is crucial for reducing such ambiguities. However, existing models inadequately leverage acceleration due to two main factors: 1) Acceleration data is noisier than rotation measurements, while IMUs provide accurate orientation estimates via Kalman filters; 2) Raw acceleration measurements fail to effectively capture continuous joint motion states, as exemplified during transitions from stationary to constant velocity, where accelerometers record peak values only at motion onset. To address these issues, we propose a two-stage model that estimates joint velocities with IMUs in the first stage, enhancing the utilization of acceleration data.

\textbf{Spatial Relationship Exploitation of Human Body Parts and Wearable Sensors}:
In IMU-based HPE, previous studies have focused on using temporal information to reconstruct complex motions, such as the ambiguity between sitting and standing or long-sitting ~\cite{yi2022physical, jiang2022transformer}. However, there has been less exploration into the varied distributions within IMU-based HPE datasets. For instance, certain upper limb motions are exclusively associated with standing in the training data, leading to potential mistakenly reconstructed in scenarios where similar motions occur while sitting during testing. To tackle this issue, our model draws inspiration from two key observations noted in recent research~\cite{lee2021uncertainty, shuai2022adaptive, shuai2022adaptive}: 1) globally rare poses often comprise local joint configurations that are frequently represented in training datasets; 2) there is a significant dependency among nearby joints, which decreases as the distance between joints increases.
Considering these insights, we propose a part-based HPE model that divides the human body into three regions: upper limbs, torso, and lower limbs, which allows the model to concentrate on the distinct characteristics of different body regions. By acknowledging the spatial relationships of body parts and sensor distribution, our model aims to enhance accuracy in pose estimation, especially in scenarios where similar motions are performed in different postures.



\textbf{Human Pose Estimation Model Generalization}: 
The scarcity of motion capture data with inertial measurement in the Skinned Multi-Person Linear (SMPL) model format~\cite{loper2023smpl} restricts the performance of IMU-based models. To address this, earlier research, beginning with DIP~\cite{huang2018deep}, utilized virtual IMU data generated from the AMASS dataset~\cite{mahmood2019amass} to increase the diversity of training samples. While this virtual-to-reality approach has shown efficacy in various domains~\cite{pei2021mars, xia2021learning, ma20233d, peng2023source}, a notable discrepancy remains between virtual and real IMU measurements, hindering further advancements in IMU-based HPE tasks.

As human pose estimation research gains traction, the availability of motion capture datasets with real IMU measurements is increasing. However, a significant challenge in utilizing these real datasets is the variation in skeleton formats they present. Our work introduces a straightforward yet efficient mapping strategy to reconcile different skeleton formats, allowing the incorporation of additional real-world motion capture datasets with actual IMU measurements into the training of IMU-based HPE models. This integration results in improved accuracy and generalization in pose estimation. The contributions of this paper are summarized as follows:
\begin{itemize}
    \item Our research introduces an innovative two-stage deep learning model designed for real-time and robust human pose estimation utilizing sparse IMU sensors, called DynaIP (Dynamic Inertial Poser). This method uniquely addresses motion ambiguity by learning pseudo velocities, which allows for the full utilization of acceleration data and leverage the strengths of sparse sensor data. 
    
    \item Our approach divides the human body and associated IMU sensors into three local regions, forming the basis for our part-based human pose estimation model. This model incorporates low-dimensional global motion information to avoid the full-body pose inconsistencies. By focusing on individual body parts, the model minimizes the influence of less associated joints, thereby enhancing the robustness and reliability of motion tracking.
    
    

    \item We incorporate more real-world Mocap data with IMU measurements across different skeleton formats and applied them uniformly in IMU-based HPE model training. The extensive experimental results demonstrate that our approach significantly outperforms competitors and exhibits good generalization performance.

\end{itemize}

\section{Related Work}
\label{sec:Related Work}

\begin{figure*}[h]
	\centering
	\includegraphics[width=17.5cm]{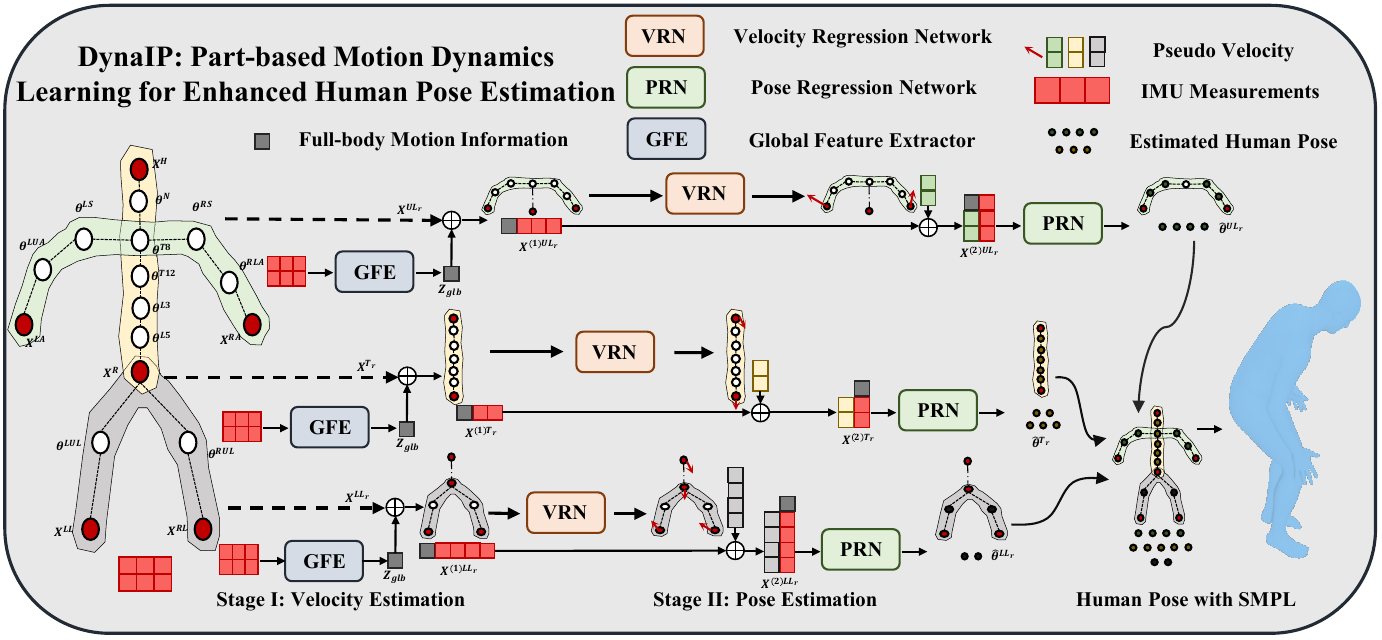}
	\caption{{ Overview of our proposed method, a part-based human pose estimation model with pseudo-velocity regression. Our model incorporates a two-stage structure. The first stage predicts joint velocities using IMU measurements, while the second stage focuses on predicting the entire body's joints rotation. Additionally, we partition the human body and the attached IMU sensors into three local regions. These regions are input into our proposed part-based human pose estimation model, designed to estimate each local region's pose while maintaining global coherency. This multi-stage and part-based approach enhances the accuracy and consistency of our pose estimation.}}
	\label{overview}
	\vspace{-3mm}
\end{figure*}

HPE has been widely explored using different methods, including visual~\cite{kocabas2020vibe, geng2021bottom, tripathi20233d, shen2023learning}, inertial~\cite{huang2018deep, kaufmann2021pose, yi2022physical}, wireless~\cite{chen2022mmbody, xue2022m4esh}, and various hybrid approaches~\cite{ren2023lidar, zhang2020fusing}. 
Our paper specifically concentrates on IMU-based human pose estimation solutions, delving into their unique advantages and potential applications.

\textbf{Human Pose Estimation with IMU sensors.} 
Inertial motion capture (mocap) systems, known for their freedom from occlusion and lighting limitations, have witnessed significant progress in recent years. Commercial systems like Xsens~\cite{schepers2018xsens} are accurate but require numerous sensors, making them less practical. Early attempts~\cite{slyper2008action, Tautges_2011} to address sensor noise and observation gaps in sparse setups involved reconstructing human motion from sparse accelerometers by referencing pre-recorded databases.
Subsequent work by Marcard et al.~\cite{von2017sparse} introduced an offline optimization method using the SMPL model to fit sparse IMU data. Huang et al.~\cite{huang2018deep} made strides by applying deep learning to real-time regression of SMPL pose parameters from IMUs, though they did not fully leverage acceleration information. Yi et al.~\cite{yi2021transpose, yi2022physical} improved upon this with a multi-stage birnn structure, hierarchically regressing joint locations and integrating a physical optimizer. Jiang et al.~\cite{jiang2022transformer} introduced stationary body points (SBP) as an additional training target, utilizing zero-velocity information to address motion drift.
However, existing methods have not effectively harnessed the multi-modal information within IMUs and the spatial information of the human body. In our paper, we divide the human body and the worn IMU sensors into three regions and devise a two-stage structure with three part-based branches for human pose estimation.

\textbf{Handling Data Scarcity in Inertial Motion Capture.} 
Large-scale training data plays a crucial role in the development of learning-based methods. In previous IMU-based research, the SMPL model has been commonly used to represent human pose. However, datasets that provide SMPL ground truth, such as those used in works like~\cite{huang2018deep, trumble2017total}, are limited in size and diversity. The acquisition of SMPL ground truth data, whether through marker-based systems with the Mosh++~\cite{loper2014mosh} algorithm or offline optimizations, is prohibitively expensive, which has restricted dataset availability.
To address this limitation, IMU-based methods like~\cite{huang2018deep, yi2021transpose, jiang2022transformer, yi2022physical} have predominantly utilized the AMASS dataset to generate a wealth of virtual sensor data. They have employed a virtual-to-reality transfer learning approach during model training. While virtual IMU data has been valuable in augmenting training samples, there remains a noticeable gap between virtual and real measurement noise. This gap has hindered further improvements in performance~\cite{kwon2020imutube, xia2021learning}.
In contrast to SMPL pose parameters, ground truth obtained through mocap systems with their native skeleton representation is more convenient. There are datasets available that provide inertial mocap data with Xsens ground truth, such as~\cite{guidolin2022unipd, maurice2019human, palermo2022raw, geissinger2020motion, emokine}. 
However, it's worth noting that these datasets use different skeleton formats for their ground truth. For example, the skeleton formats of SMPL and Xsens differ, shown in Fig.~\ref{introduction}. While vision-based solutions have addressed this issue by combining multiple datasets with varying ground truth representations and using novel autoencoders to mitigate positional disparities in ground truth~\cite{sarandi2023learning}, our approach takes a simpler yet effective route. We introduce a mapping strategy that enables us to seamlessly incorporate additional real-world MoCap datasets with different skeleton formats into our training process. This strategy allows us to harness the rich diversity of available data for improved performance.

\section{Method}

Our objective is to accurately estimate human pose, denoted as $\boldsymbol{\hat{\theta}}$, using data from six sensors worn on different body parts. The corresponding IMU data can be represented by $\boldsymbol{X} \in \mathcal{R}^{T \times 72} = [\boldsymbol{X^R}, \boldsymbol{X^{LL}}, \boldsymbol{X^{RL}}, \boldsymbol{X^{H}}, \boldsymbol{X^{LA}}, \boldsymbol{X^{RA}}]$, marked in Fig.~\ref{overview}, where $T$ is the length of data samples. 
Additionally, our model predicts the pseudo velocity ($\boldsymbol{\hat{V}}$) of the joints with IMU measurements. As shown in Fig.~\ref{overview}, our model comprises three modules: 1) A method for unifying training data across different skeleton formats using a global orientation mapping strategy. 2) A two-stage human pose estimation structure with pseudo velocity regression. 3) A part-based 3D human dynamics learning module with low-dimensional full-body motion information. 

\subsection{Training Data Unified across Skeleton Formats}
\begin{figure}[!t]
	\centering
	\includegraphics[width=7.5cm]{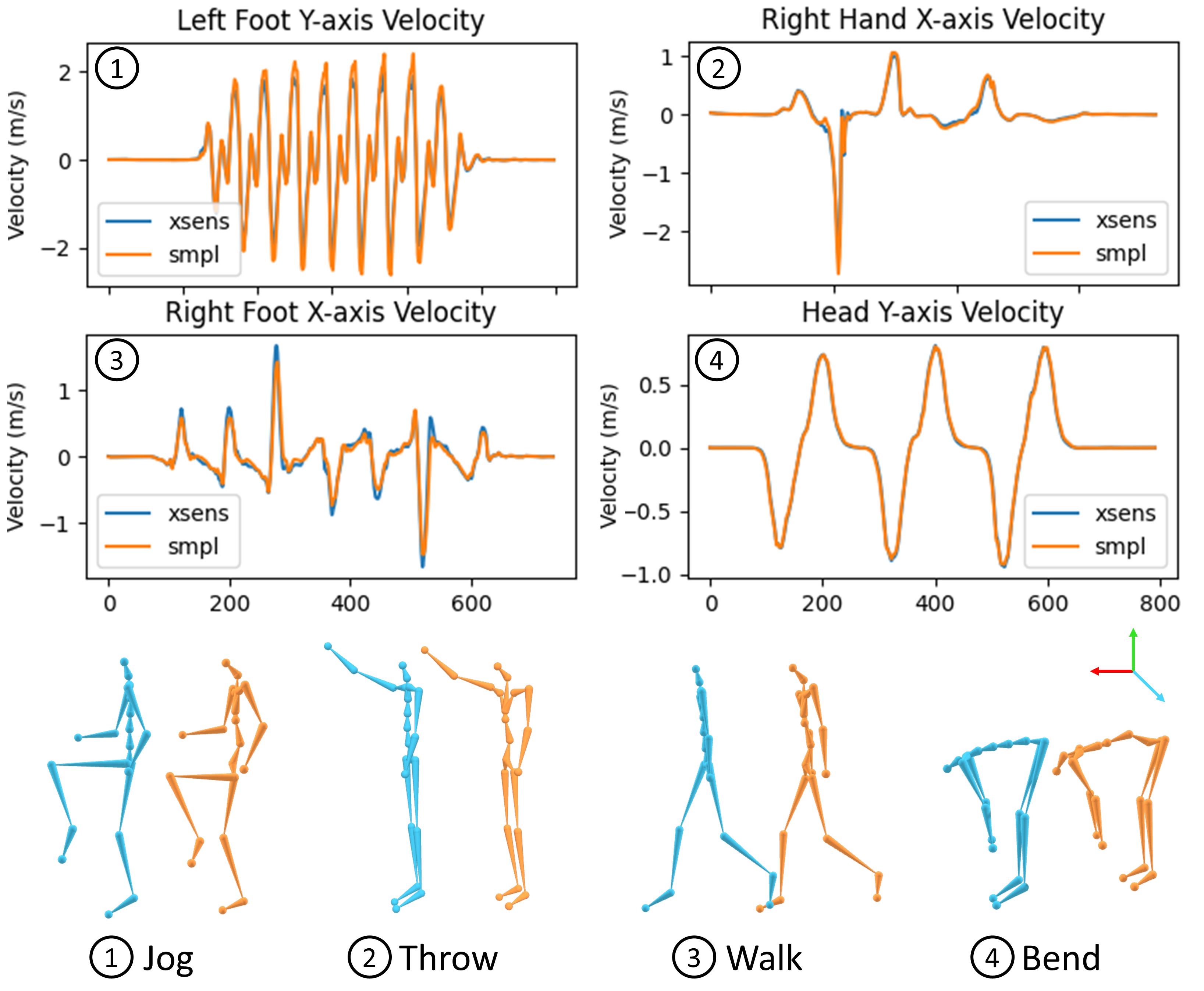}
	\caption{{ The evaluation of the end-effector velocity across various motions. With the mapping of joint's global orientations from Xsens to SMPL, there is no significant discrepancy in the end-effector velocities. }}
	\label{fig:mapping_vel}
	\vspace{-5mm}
\end{figure}

Early learning-based methods~\cite{huang2018deep, yi2021transpose} for IMU-based HPE encountered a significant challenge: the lack of datasets that contain real IMU data alongside corresponding ground truths. To address this challenge, these methods have taken an innovative approach by generating virtual IMU data using the AMASS dataset~\cite{mahmood2019amass}, improving the model's generalization capabilities due to its diversity. Fine-tuning the model with a small amount of collected real data resulted in impressive pose estimation results. However, recent studies~\cite{kwon2020imutube, jiang2022transformer, xia2021learning} have highlighted a performance degradation issue stemming from differences in noise distributions between virtual and real IMU data. This issue has been discussed in various tasks~\cite{zhang2022transfer, hoyer2022daformer}, indicating the need for a solution. As interest in IMU-based HPE grows, several datasets containing real IMU measurements have emerged~\cite{guidolin2022unipd, maurice2019human, palermo2022raw, geissinger2020motion}. These datasets represent human pose ground truths but use different skeleton formats. Instead of solely addressing the gap between virtual and real domains~\cite{kwon2020imutube, huang2018deep}, our approach takes a novel direction. We propose a training method that initially integrates motion capture datasets with real IMU data from various skeleton formats, offering a promising solution to enhance pose estimation accuracy.

The SMPL~\cite{loper2023smpl} and Xsens~\cite{schepers2018xsens} skeletons are two widely used skeleton representations. While they share overall structural similarities, one notable difference lies in the number of torso joints, as depicted in Fig.~\ref{introduction}. Previous research efforts~\cite{dai2022hsc4d, guzov2022visually} have achieved success in mapping human poses captured using commercial inertial mocap systems onto the SMPL skeleton. This mapping process involves replicating relative joint rotations and excluding redundant torso joints. 
In the skeleton model, each 3D rigid bone's motion can be denoted by a homogeneous matrix $M_{bone}\in SE(3)$~\cite{murray2017mathematical}.
\begin{equation}
\small
    M_{bone} =\begin{bmatrix}
      \boldsymbol{R_{b}} & \boldsymbol{p_{b}}\\
      \boldsymbol{0}_{1\times 3}& 1
    \end{bmatrix},  
    \vspace{-1mm}
\end{equation}
where $\boldsymbol{R_{b}}\in SO(3)$ is a global rotation matrix and $ \boldsymbol{p_{b}}\in \mathcal{R}^3$ is a predefined bone displacement. For inertial-based mocap, IMUs, attached to the human body, offer sensor's global orientation $\boldsymbol{R_{s}}\in SO(3)$. After calibration~\cite{huang2018deep, yi2021transpose, yi2022physical}, the sensor orientation $\boldsymbol{R_{s}}$ serves as a direct representation of the respective bone's orientation $\boldsymbol{R_{b}}$.
As such, although specific bone displacements $\boldsymbol{p_{bone}}$ can be defined differently within SMPL and Xsens skeleton, the bone orientations would be consistent with the IMU readings because they correspond to the same location on human body. Therefore, we've established a one-to-one ground truth mapping across skeletons, e.g., wrists to wrists, legs to legs, and so on. For the torso, we eliminate this redundant joint to maintain the consistency of our mapping. With this mapping strategy, we could unify the inertial mocap data across different skeleton formats.

In line with previous tasks related to human motion retargeting~\cite{aberman2020skeleton, mourot2023humor}, we utilize the velocity of end-effectors as a qualitative measure to demonstrate the practicality of our mapping process. As depicted in Fig.~\ref{fig:mapping_vel}, our global orientation mapping strategy consistently maintains accuracy across various motions. This serves as a robust foundation for unifying a more extensive dataset of real IMU mocap data. By training the model with this unified dataset, we enable it to fully leverage real IMU data, resulting in improved generalization performance compared to the traditional virtual-to-reality approach. 

\subsection{Two-stage Human pose estimation with Pseudo Velocity Regression}
Given our approach's reliance on sparse IMUs attached to limb ends for estimating full-body pose, there is an inherent challenge in accurately estimating joint rotations without direct IMU measurements. 
Previous models~\cite{huang2018deep, jiang2022transformer} often overlook the rich multi-modal information contained in the IMUs, primarily relying on global orientation measurements. Therefore, it is essential to effectively integrate the dynamic information provided by acceleration into our learning-based method. To address this, we introduce a two-stage HPE model that estimates, in two distinct stages, the velocity of joints with IMUs and the global rotation of each joint in the full-body. Our method comprises two primary stages: Pseudo velocity regression (Stage I) and human pose estimation (Stage II). This two-stage approach allows us to leverage both acceleration and orientation data to enhance pose estimation accuracy.


\subsubsection{Stage I: pseudo velocity regression}

As previously mentioned in Sec.~\ref{sec:intro}, relying on raw acceleration data may not effectively capture the continuous motion states of joints due to its sensitivity to instantaneous dynamics. In response to this challenge, Stage I of our model is dedicated to regressing the pseudo velocity of the joints with IMUs. This step is crucial for effectively extracting the dynamic information contained in acceleration data. The underlying principles guiding this design are twofold:
\begin{itemize}
    \item \textbf{Velocity as a Motion Indicator:} Velocity has been demonstrated to better reflect the joints motion dynamics, as supported by previous research~\cite{yang2021lobstr, herath2020ronin}. It plays a crucial role in compensating for the motion ambiguity that can arise from underutilizing different modalities.
    \item \textbf{Velocity Estimation:} Velocity can be obtained by integrating acceleration and ensure that the dynamic information remains accurate and reliable, unlike raw acceleration. We employ a neural network to estimate the velocities of joints with IMUs, thereby harnessing the IMU's potential to express continuous body dynamics.
\end{itemize}

Our model's Stage-I takes the raw IMU measurements $\boldsymbol X = [\boldsymbol{X^R}, \boldsymbol{X^{LL}}, \boldsymbol{X^{RL}}, \boldsymbol{X^{H}}, \boldsymbol{X^{LA}}, \boldsymbol{X^{RA}}]$ as input, the velocity $\boldsymbol{\hat{V}}=[\boldsymbol{\hat{V}^R}, \boldsymbol{\hat{V}^{LL}}, \boldsymbol{\hat{V}^{RL}}, \boldsymbol{\hat{V}^{H}}, \boldsymbol{\hat{V}^{LA}}, \boldsymbol{\hat{V}^{RA}}]$ as output via the model's Velocity Regression Network (VRN) module, denoted by $S_{VRN}(\cdot)$. Following the advanced learning-based RNN initialization strategy proposed by PIP~\cite{yi2022physical}, our model additionally takes the initial velocities$\boldsymbol {V_0}$ of leaf joints as the model's input. After that, we get the predicted pseudo velocity $\boldsymbol {\hat{V}}$, represented by
\begin{equation}  \small
	\boldsymbol {\hat{V}} = S_{VRN}(\boldsymbol{X^{(1)}}, \boldsymbol V_0),
 \vspace{-1mm}
\end{equation}
where $\boldsymbol{X^{(1)}} = \boldsymbol X$ represents the input of the first stage.

The VRN module is comprised of two main components: a Multi-Layer Perceptron (MLP) and a two-layer Long Short-Term Memory (LSTM) network. The MLP takes the initial velocity as input. Its primary purpose is to process and transform this initial velocity information. The outputs generated by the MLP are then assigned to serve as the hidden state and cell state inputs for the first frame of the LSTM network. This design allows for the initial velocity information to be effectively processed and used as a starting point for the LSTM, which can then continue to capture and predict the velocity and motion dynamics over subsequent frames. For the intermediate variable $\boldsymbol {\hat{V}}$ to be supervised, we obtain the velocity ground-truth $\boldsymbol V =( FK(\boldsymbol{\theta_t}) - FK(\boldsymbol{\theta_{t-1}})) / \Delta t$ through the ground-truth of human pose  $\boldsymbol \theta$ and forward kinematics $FK(\cdot)$ for supervision. The pseudo-velocity loss $L_{vel}$ could be represented by:
\begin{equation}
\small
\begin{aligned}
& \label{Lvel}
	L_{vel} = \|\boldsymbol V - \boldsymbol {\hat{V}} \|_2 \\
	&= \|\frac{FK(\boldsymbol{\theta_t}) - FK(\boldsymbol{\theta_{t-1}})}{\Delta t} -  S_{VRN}(\boldsymbol X, \boldsymbol V_0) \|_2,
\end{aligned}%
\end{equation}
where $\Delta t$ represents the time interval and $\boldsymbol{\theta_t}$ the ground-truth of human pose at time $t$. For the velocity of the root joint, we only preserve its vertical component. 

\subsubsection{Stage II: human pose estimation}
Building upon the dynamic information extracted by the VRN module, our model proceeds to Stage II, where the primary goal is to achieve robust human pose estimation. 
For Stage II, we design a Pose Regression Network (PRN) module, denoted by $S_{PRN}(\cdot)$, to estimate the joint rotations $\boldsymbol{\hat{\theta}}$ from the IMU measurements $\boldsymbol X$ and estimated leaf joints velocities $\boldsymbol{\hat{V}}$, which are concatenated as input $\boldsymbol{X^{(2)}} = [\boldsymbol{\hat{V}}, \boldsymbol X]$ to the Stage II. Based on the initialization strategy and the PRN module, our model could output the estimated human pose $\boldsymbol{\hat{\theta}}$:
\begin{equation}  \small
	\boldsymbol {\hat{\theta}} = S_{PRN}(\boldsymbol{X^{(2)}}, \boldsymbol{\theta_0}),
\end{equation}
where the $\boldsymbol{\theta_0}$ is the initial human pose, and the PRN $S_{PRN}(\cdot)$ is applied with a two-layers LSTM. 

With the predicted human pose $\boldsymbol{\hat{\theta}}$ and the ground-truth $\boldsymbol{\theta}$, we could get the loss $L_{pose}$ represented by:
\begin{equation}
\small
\begin{aligned}
\label{Lpose}
	L_{pose} = \|\boldsymbol{\theta} - \boldsymbol{\hat{\theta}} \|_2 = \|\boldsymbol{\theta} - S_{PRN}(\boldsymbol{X^{(2)}}, \boldsymbol{\theta_0}) \|_2,
\end{aligned}%
\end{equation}
Our two-stage network, which includes velocity regression, successfully addresses the ambiguities arising from the under-utilization of acceleration. However, considering the distinctive motion patterns of the upper and lower limbs, as well as the infrequent motion combinations, it is crucial to harness the spatial information of the human body to ensure robust motion tracking.


\subsection{Learning Part-based 3D Human Dynamics with Three Local Body Regions}
Directly using all six IMU measurements as the model's input, as done in previous approaches~\cite{huang2018deep, yi2022physical}, without considering the inherent spatial relationship of the human body, can lead to motion ambiguity due to weak associations between body parts.
In this section, drawing inspiration from~\cite{lee2021uncertainty, shuai2022adaptive}, we introduce local region modeling to mitigate this issue. Our model divides the entire body into three local regions: upper limbs region ($UL_r$), torso region ($T_r$), and lower limbs region ($LL_r$), as shown in Fig.~\ref{overview}.

Correspondingly, the IMU measurements and the estimated human pose in our proposed model are partitioned into three local regions. To create the model's input, we group the IMU sensors on our bodies into three sets, denoted as ${{\boldsymbol{{X^l}}} , {l \in \left\{ {{{UL_r, T_r, LL_r}}} \right\}}}$, which can be represented by $\boldsymbol{X^{UL_r}} = [\boldsymbol{X^{R}}, \boldsymbol{X^{LA}}, \boldsymbol{X^{RA}}], \boldsymbol{X^{T_r}} = [\boldsymbol{X^R}, \boldsymbol{X^H}], \boldsymbol{X^{LL_r}} = [\boldsymbol{X^R}, \boldsymbol{X^{LL}}, \boldsymbol{X^{RL}}, \boldsymbol{X^H}]$.

Following the previously mentioned two-stage human pose estimation structure, we establish three sub-models to acquire the three groups of inputs, and estimate the part-based pseudo velocity $\boldsymbol{\hat{V^{l}}}, l \in \{UL_r,T_r,LL_r\}$ and the joint rotation outputs $\boldsymbol{\hat{\theta^{l}}}, l \in \{UL_r,T_r,LL_r\}$. Our estimated joints of full-body are also output in three sub-modules, specifically expressed as $\boldsymbol{\theta^{UL_r}} = [\boldsymbol{\theta^{LS}}, \boldsymbol{\theta^{LUA}}, \boldsymbol{\theta^{RS}}$, $\boldsymbol{\theta^{RUA}}], \boldsymbol{\theta^{T_r}} = [\boldsymbol{\theta^{LUL}}, \boldsymbol{\theta^{RUL}}]$, $\boldsymbol{\theta^{LL_r}} = [\boldsymbol{\theta^{L5}}, \boldsymbol{\theta^{L3}}, \boldsymbol{\theta^{T12}}, \boldsymbol{\theta^{T8}}, \boldsymbol{\theta^{N}}]$. 
The representation of the joints is shown in Fig.~\ref{overview}.


With our part-based two-stage structure, our model achieves a comprehensive representation of full-body joint rotations by synthesizing the outputs from three sub-models. This part-based design, relying on local inputs and features, effectively learns unique pose configurations for each body part and minimizes the negative impact of weakly associated joints. However, one challenge we face with this part-based approach is the potential lack of global coherence in the estimated poses. Furthermore, to address this, we incorporate global body motion information into our part-based HPE model, drawing inspiration from previous work~\cite{lee2021uncertainty, zeng2020srnet}. Our model utilizes a global feature extractor, denoted as $S_{GLB}$, to coarsely capture the full-body motion information $\boldsymbol{Z_{glb}}$ from all six IMUs $\boldsymbol X$. Subsequently, the global representation is concatenated with the inputs of the two-stage network featuring three local region branches. The input of Stage I ($\boldsymbol{X^{(1)_{l}}}$) and Stage II ($\boldsymbol{X^{(2)_l}}$) could be re-represented by 
\begin{equation}
\small
\begin{aligned}
& \label{Two-Input}
    \boldsymbol{X^{(1)_l}} = [\boldsymbol {X^l}, \boldsymbol{Z_{glb}}], \\
	& \boldsymbol{X^{(2)_l}} = [\boldsymbol {\hat{V}^l}, \boldsymbol {X^l}, \boldsymbol{Z_{glb}}], l \in \{UL_r,T_r,LL_r\},
\end{aligned}%
\vspace{-2mm}
\end{equation}
where the $\boldsymbol {\hat{V}^l} = S^{l}_{VRN}(\boldsymbol{X^{(1)_l}}, \boldsymbol V^l_0)$ is the local pseudo velocity estimated by the local VRN module $S^{l}_{VRN}(\cdot)$, and $\boldsymbol V^l_0$ means the initial velocities of the local regions.
Therefore, we get the loss function for two stages:
\begin{equation}
\small
\begin{aligned}
& \label{Two-Input}
    L^l_{vel} = \|\boldsymbol V^l - \boldsymbol {\hat{V}^l} \|_2, L^l_{pose} = \|\boldsymbol{\theta^l} - \boldsymbol{\hat{\theta^l}} \|_2,
\end{aligned}%
\vspace{-2mm}
\end{equation}
where the $\boldsymbol V^l$, $\boldsymbol{\theta^l}, l \in \{UL_r, T_r, LL_r\}$ are the ground-truth of the velocity and human pose for three local regions, and the $\boldsymbol {\theta^l}$ is the local human pose estimated by the local PRN module $S^{l}_{PRN}(\boldsymbol{X^{(2)_l}}, \boldsymbol {\theta^l_0})$ of three local branches. 
By integrating local regions with global information, our model ensures that while each part-based branch effectively learns localized motion dynamics, the overall pose estimation remains coherent and consistent with the global motion patterns of the human body. This approach strikes a balance between local and global information, resulting in robust and accurate pose estimation.

With the technical analysis provided above, we can formulate the final objective function for training our model as follows:
\begin{equation}  
\small
	L = \sum_{l \in \{UL_r,T_r,LL_r\}} (L^l_{pose} + L^l_{vel})
\end{equation}


\section{Experiments}
\label{sec:experiments}
\textbf{Experiment Setup.} Our experimental evaluation is structured into three main parts: Firstly, we illustrate the benefits of using unified inertial mocap data compared with a virtual-to-real training scheme. Secondly, we show the effectiveness of our proposed model with other state-of-the-art methods on our unified mocap data. Finally, we conduct ablation studies on key components of our model.

\textbf{Datasets.} We utilized a combination of datasets for both training and evaluation in our experiments. The datasets include DIP-IMU~\cite{huang2018deep} and Xsens datasets, which encompass AnDy~\cite{maurice2019human}, Emokine~\cite{emokine}, Virginia Natural Motion~\cite{geissinger2020motion}, UNIPD~\cite{guidolin2022unipd}, and CIP~\cite{palermo2022raw}. Additionally, we used the AMASS~\cite{mahmood2019amass} dataset for evaluating the effectiveness of virtual and real mocap data. Detailed information about these datasets can be found in the supplementary document.

\textbf{Metrics.} Following~\cite{yi2021transpose, yi2022physical}, we use the following metrics for evaluation:
\textit{1) SIP error [°]}: the mean global rotation difference of upper arms and upper legs between the estimation and ground truth; \textit{2) Global Angular error [°]}: the mean global rotation error between estimated joints and ground truth; \textit{3) Position error [cm]}: the mean Euclidean distance error between all joints and ground truth, root position is aligned. \textit{4) Mesh error [cm]}: the mean vertex distance between estimated meshes and ground truth.

\subsection{Impact of the Unified Inertial Mocap Data and Virtual-to-Real Training Scheme}
\textbf{Comparing performance of our model using different training settings.} To evaluate the impact of unified training data and virtual dataset, we conducted an experiment on these four training settings: \textit{1)} Synthetic AMASS data only; \textit{2)} Synthetic AMASS data with DIP-IMU Fine-tuning; \textit{3)} Real Xsens data only (DynaIP); \textit{4)} Combined DIP-IMU and Xsens data (DynaIP*). With these four settings, we evaluated our model on the DIP-IMU test set. Furthermore, for models only trained on AMASS or Xsens data, additional evaluations were conducted on the full DIP-IMU dataset (includes the training set defined by the previous methods) and specifically on DIP-IMU challenging sitting sequences. 

The experiment results are presented in Tab.~\ref{table:real_virtual_our}. Comparing the performance of models \textit{1)} and \textit{3)} on the DIP-IMU full set, it becomes evident that while synthetic data from AMASS demonstrates some transferability, real mocap data from Xsens exhibits significantly better performance and generalization ability, resulting in a 22\% reduction in the SIP error. This improvement is particularly pronounced in the DIP-IMU sitting sequence, where there is a 28\% reduction in the SIP error. Additionally, when the DIP training set is integrated into our model training, two noteworthy observations emerge: First, the inclusion of the DIP training set enhances the performance of models \textit{1)} and \textit{3)}. Second, a comparison between models \textit{2)} and \textit{4)} reveals a clear advantage in using a more comprehensive unified real mocap dataset over the virtual-to-real scheme. The additional diversity and realism provided by the real mocap data contribute to enhancing the model's generalization capabilities.

\begin{table}
\centering
\resizebox{0.45\textwidth}{!}{
\begin{tabular}{lccccc}
\toprule
\multirow{2}{*}{} & \multicolumn{4}{c}{DIP IMU Test} & \\
\cmidrule{2-5}
 & SIP Err(°) & Ang Err(°) & Pos Err(cm) & Mesh Err(cm) \\
\midrule
\textit{1)}AMASS & 23.80 & 8.25 & 6.04 & 7.18\\
\textit{2)}AMASS+DIP & 14.41 & 5.90 & 5.03 & 6.05\\
\textit{3)}Xsens(DynaIP) & 17.31 & 7.66 & 5.79 & 7.01 \\
\textit{4)}Xsens+DIP(DynaIP*) & \textbf{13.67} & \textbf{5.83} & \textbf{4.84} & \textbf{5.82}\\
\midrule
\multirow{2}{*}{} & \multicolumn{4}{c}{DIP IMU Full} & \\
\cmidrule{2-5}
 & SIP Err(°) & Ang Err(°) & Pos Err(cm) & Mesh Err(cm) \\
\midrule
\textit{1)}AMASS & 24.50 & 8.16 & 6.87 & 7.82\\
\textit{3)}Xsens & \textbf{18.98} & \textbf{7.85} & \textbf{6.73} & \textbf{7.80} \\
\midrule
\multirow{2}{*}{} & \multicolumn{4}{c}{DIP IMU Sitting} & \\
\cmidrule{2-5}
 & SIP Err(°) & Ang Err(°) & Pos Err(cm) & Mesh Err(cm)\\
\midrule
\textit{1)}AMASS & 40.72 & 11.98 & 13.85 & 14.56 \\
\textit{3)}Xsens & \textbf{29.28} & \textbf{10.76} & \textbf{12.52} & \textbf{13.60} \\
\bottomrule
\end{tabular}
}
\caption{The performance comparison with different training data settings on our model.}
\label{table:real_virtual_our}
\vspace{-3mm}
\end{table}

\begin{figure}[!t]
	\includegraphics[width=8cm]{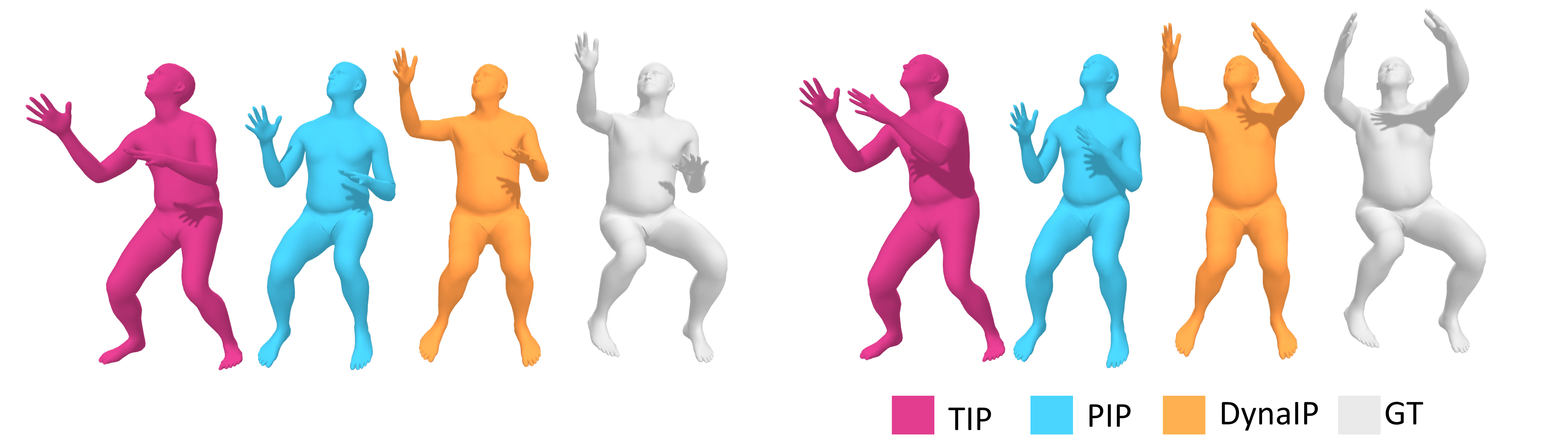}
	\caption{Qualitative comparisons on DIP-IMU~\cite{huang2018deep} test set.}
	
	\label{fig:dip_qualitative}
	\vspace{-4mm}
\end{figure}

\textbf{Comparing performance of our method and previous virtual-to-real SOTAs.} We further compare the results trained on the unified real mocap dataset with previous virtual-to-real transfer approaches.
In Tab.~\ref{table:DIP_sota}, when trained with the DIP-IMU training set, DynaIP* demonstrates an 8\% reduction in SIP error and a 19\% reduction in global pose error compared to PIP~\cite{yi2022physical}, highlighting the accuracy and robustness of our approach. Furthermore, DynaIP, which only uses Xsens data, achieves comparable performance to TransPose~\cite{yi2021transpose}, which employs DIP-IMU for fine-tuning. This illustrates that having access to abundant real inertial mocap data is beneficial for generalizing to new subjects and unconstrained motions.
Fig.~\ref{fig:dip_qualitative} provides a visual comparison of our model's performance on the DIP-IMU test set against state-of-the-art methods trained with the AMASS and DIP-IMU. We specifically focus on challenging poses within the DIP-IMU test set, particularly hand-raising motions with relatively low velocities. In these scenarios, our method shows superior performance, effectively capturing the nuances of these motions where the utilization of acceleration is crucial, highlighting the excellence of subtle motion dynamics understanding.

\begin{table}
\centering
\resizebox{0.4\textwidth}{!}{
\begin{tabular}{lcccc}
\toprule
\multirow{2}{*}{} & \multicolumn{4}{c}{DIP IMU Test} \\
\cmidrule{2-5}
 & SIP Err(°) & Ang Err(°) & Pos Err(cm) & Mesh Err(cm)\\
\midrule
DIP~\cite{huang2018deep} & 17.10 & 15.16 & 7.33 & 8.96\\
TransPose~\cite{yi2021transpose} & 16.68 & 8.85 & 5.95 & 7.09\\
TIP~\cite{jiang2022transformer} & 16.20 & 9.17 & 5.49 & 6.61\\
PIP~\cite{yi2022physical} & 15.02 & 8.73 & 5.04 & \textbf{5.95}\\
\midrule
DynaIP & 17.43 & 8.90 & 5.93 & 7.71\\
DynaIP* & \textbf{13.78} & \textbf{7.07} & \textbf{4.98} & 5.99\\
\bottomrule
\end{tabular}
}
\caption{The performance comparison between our model with the SOTAs reported in their papers on DIP-IMU~\cite{huang2018deep} test set. For a fair comparison, we transform each result into a local rotation representation, consistent with the PIP~\cite{yi2022physical}.}
\label{table:DIP_sota}
\vspace{-5mm}
\end{table}

\begin{figure}[!t]
\centering
	\includegraphics[width=8cm]{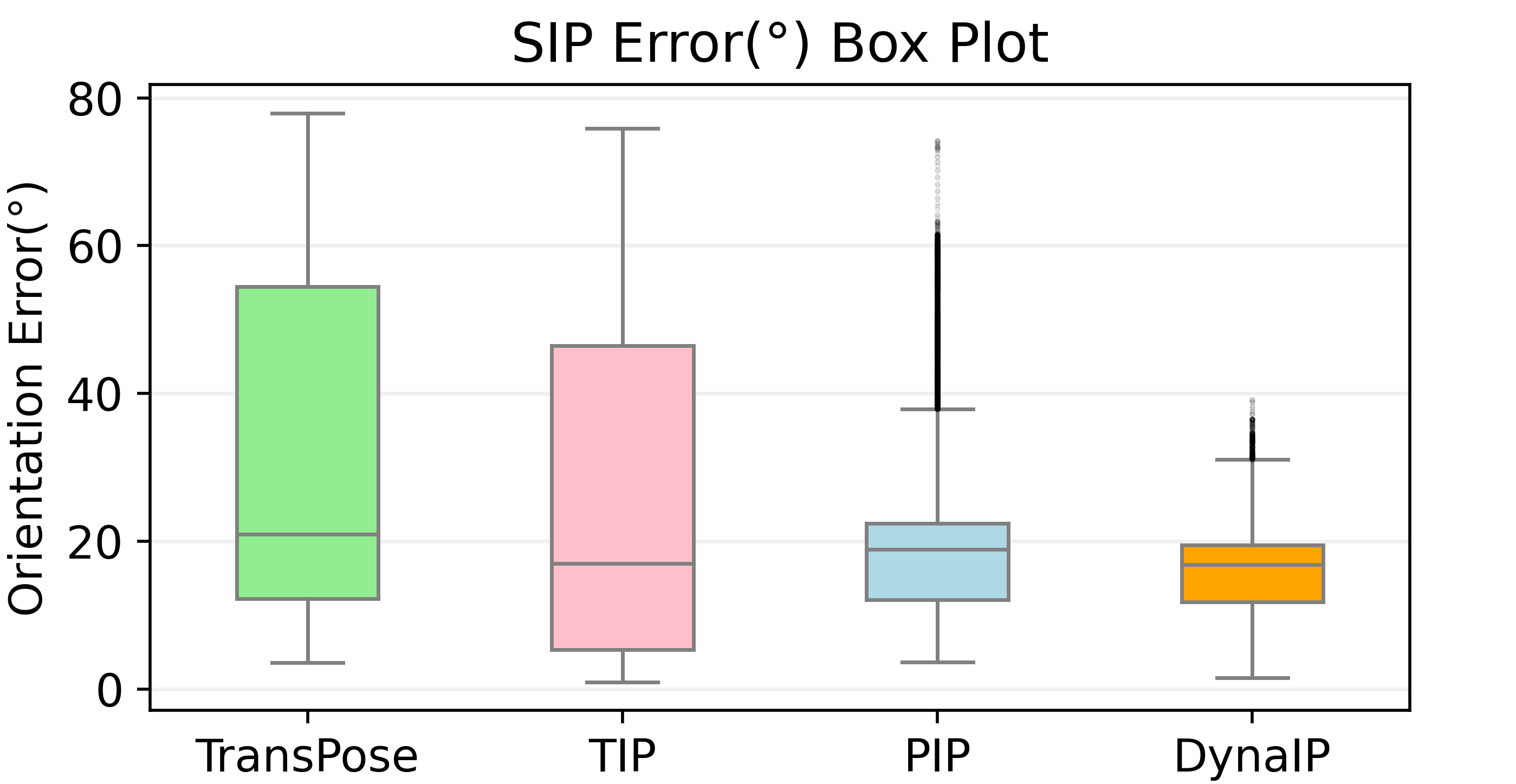}
	\caption{Qualitative results of SIP error box plot for three competing methods and DynaIP on Natural Motion~\cite{geissinger2020motion} dataset.}
	\label{fig:box_plot}
	\vspace{-5mm}
\end{figure}

\subsection{Overall Performance Comparison on the Unified Inertial Mocap Data}

\begin{table*}
\centering
\resizebox{0.98\textwidth}{!}{
\begin{tabular}{l*{15}{c}}
\toprule
\multirow{2}{*}{} & \multicolumn{3}{c}{DIP IMU} & \multicolumn{3}{c}{AnDy} & \multicolumn{3}{c}{UNIPD} & \multicolumn{3}{c}{CIP} & \multicolumn{3}{c}{Natural Motion} \\
\cmidrule(lr){2-4} \cmidrule(lr){5-7} \cmidrule(lr){8-10} \cmidrule(lr){11-13} \cmidrule(lr){14-16}
& SIP Err(°) & Ang Err(°) & Pos Err(cm) & SIP Err(°) & Ang Err(°) & Pos Err(cm) & SIP Err(°) & Ang Err(°) & Pos Err(cm) & SIP Err(°) & Ang Err(°) & Pos Err(cm) & SIP Err(°) & Ang Err(°) & Pos Err(cm) \\
\midrule
DIP~\cite{huang2018deep} & 21.71 & 10.14 & 7.92 & 11.39 & 5.73 & 4.34 & 12.02 & 5.47 & 4.22 & 19.13 & 8.61 & 6.86 & 33.43 & 11.91 & 13.33 \\
TransPose~\cite{yi2021transpose} & 22.53 & 10.28 & 8.42 & 12.15 & 6.29 & 4.91 & 15.11 & 6.05 & 4.82 & 20.06 & 8.75 & 6.86 & 30.62 & 11.26 & 11.99 \\

TIP~\cite{jiang2022transformer} & 19.22 & 8.94 & 6.91 & 10.11 & 4.55 & 3.56 & 9.85 & 4.06 & 2.78 & 13.05 & 5.67 & 4.30 & 22.06 & 7.90 & 7.92 \\

PIP~\cite{yi2022physical} & 17.62 & 8.33 & 6.21 & 9.49 & 4.09 & \textbf{3.29} & 8.90 & 3.59 & 2.66 & 12.68 & 5.52 & 4.12 & 19.62 & 7.49 & 6.93 \\

DynaIP & \textbf{17.31} & \textbf{7.66} & \textbf{5.79} & \textbf{8.93} & \textbf{3.45} & 3.41 & \textbf{7.29} & \textbf{2.77} & \textbf{2.21} & \textbf{11.42} & \textbf{4.54} & \textbf{3.69} & \textbf{15.78} & \textbf{7.18} & \textbf{5.83} \\

\bottomrule
\end{tabular}
}
\caption{Evaluation results of the state-of-the-art models on DIP-IMU~\cite{huang2018deep}, AnDy~\cite{maurice2019human}, UNIPD~\cite{guidolin2022unipd}, CIP~\cite{palermo2022raw} and Natural Motion~\cite{geissinger2020motion} when trained only with real inertial mocap data. All models run in real-time setting.}
\label{table:sota_xsens}
\vspace{-5mm}
\end{table*}

For fair and meaningful evaluation, we retrained previous models~\cite{huang2018deep, yi2021transpose, jiang2022transformer, yi2022physical} using Xsens data and evaluated their performance on both the Xsens and DIP-IMU test sets, and compared their results to our model's performance. 

The evaluation results in Tab.~\ref{table:sota_xsens} demonstrate the superior performance of our method compared to competing methods on various datasets. Notably, our model achieves a relative 28\% reduction in SIP error on the Natural Motion dataset and an 18\% reduction in global pose error on the CIP dataset compared to the state-of-the-art structure PIP~\cite{yi2022physical}. These significant improvements in performance highlight the robustness and generalization capability of our model, which can be attributed to our effective velocity estimation strategy and part-based modeling modules.
\begin{figure}[!t]
\centering
	\includegraphics[width=7.8cm]{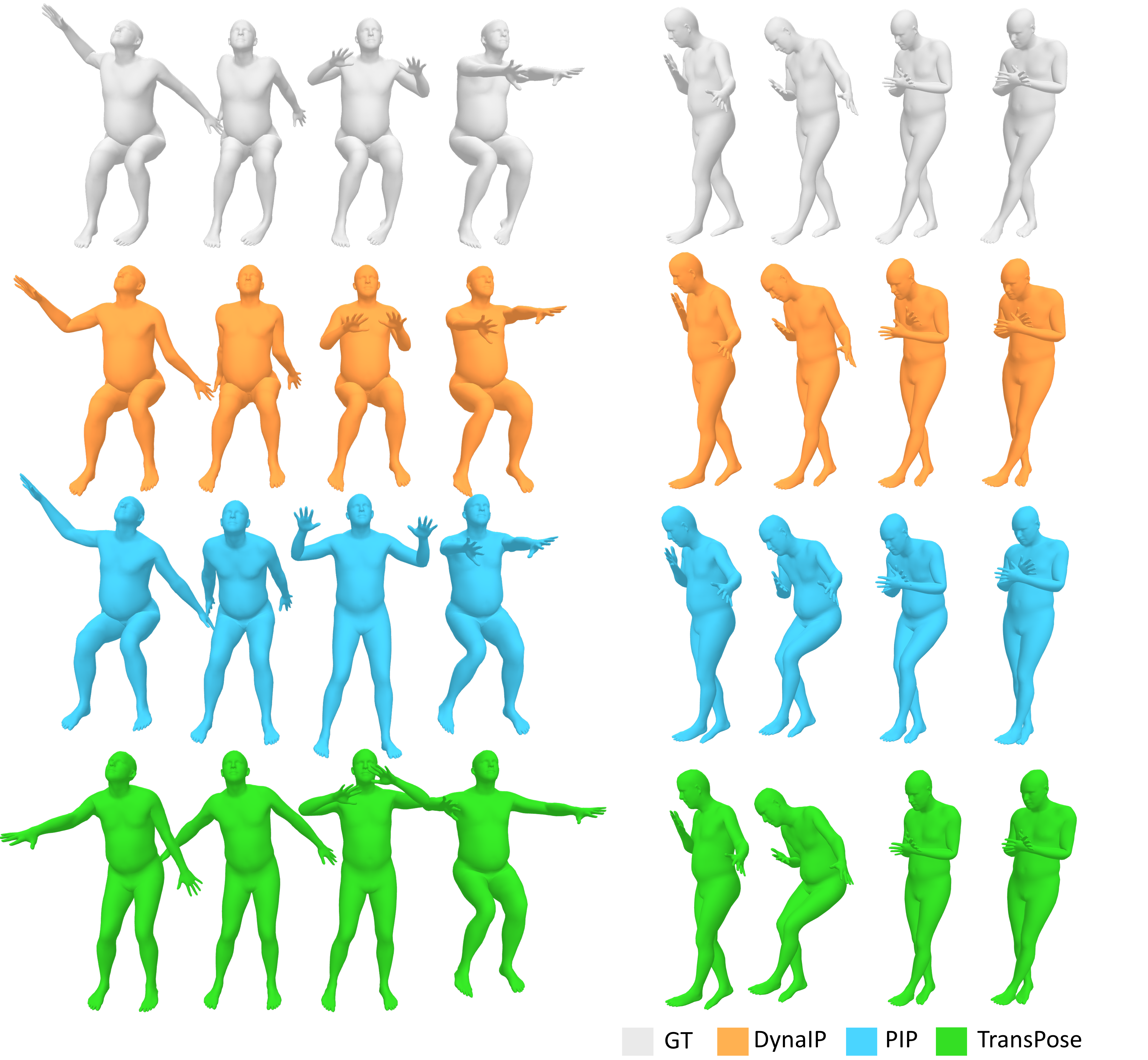}
	\caption{Qualitative comparisons of different model from CIP~\cite{palermo2022raw} (Left) and Natural Motion~\cite{geissinger2020motion} (Right) test set.}
	
	\label{fig:xsens_qualitative}
	\vspace{-3mm}
\end{figure}

Fig.~\ref{fig:box_plot} presents a box plot of the SIP error for various models~\cite{yi2021transpose, jiang2022transformer, yi2022physical} and our model on the Natural Motion dataset~\cite{geissinger2020motion}. It's worth noting that our model not only achieves the lowest maximum SIP error but also has fewer outliers, indicating that our method consistently produces more robust performance across various motions. This further emphasizes the effectiveness of our approach in handling challenging and diverse pose estimation tasks. 
In Fig.~\ref{fig:xsens_qualitative}, we provide a visualization comparison between our reconstructed poses and those inferred by state-of-the-art methods using selected frames from CIP and Natural Motion datasets.  Our model accurately maintains the sitting or standing pose, even during arm movements or prolonged periods of crossed legs. In contrast, methods like TransPose~\cite{yi2021transpose} and PIP~\cite{yi2022physical} intermittently oscillate between sitting and standing poses.
The superior performance of our model in these instances can be attributed to two factors: firstly, the retention of root vertical velocity information, which helps capture motion pattern transitions; and secondly, the effectiveness of our part-based modeling approach. This approach proves particularly advantageous for un/rare-seen poses, as it is less influenced by imbalanced training distributions and better at resolving ambiguities. 

\subsection{Ablations on the Components of our Model}

\begin{table}
\resizebox{0.46\textwidth}{!}{
\begin{tabular}{l*{6}{c}}
\toprule
& \multicolumn{2}{c}{DIP IMU} & \multicolumn{2}{c}{CIP} & \multicolumn{2}{c}{Natural Motion} \\
\cmidrule(lr){2-3} \cmidrule(lr){4-5} \cmidrule(lr){6-7}
& SIP Err(°) & Ang Err(°) & SIP Err(°) & Ang Err(°) & SIP Err(°) & Ang Err(°) \\
\midrule
\ding{172}Baseline & 15.26 & 6.17 & 13.18 & 5.13 & 33.22 & 11.20 \\
\ding{173}w/o Part & 14.97 & 6.02 & 13.00 & 4.86 & 31.62& 9.60\\
\ding{174}w/o Vel & 14.87 & 6.11 & 12.54 & 4.89 & 29.15 & 10.49\\
DynaIP* & \textbf{13.67} &\textbf{5.83} & \textbf{11.67} & \textbf{4.63} & \textbf{18.88} & \textbf{8.03} \\
\bottomrule
\end{tabular}
}
\caption{The ablation study on pseudo-velocity estimation and the part-based modeling approach.}
\label{table:ablation}
\vspace{-4mm}
\end{table}

\begin{figure}[!t]
	\includegraphics[width=8cm]{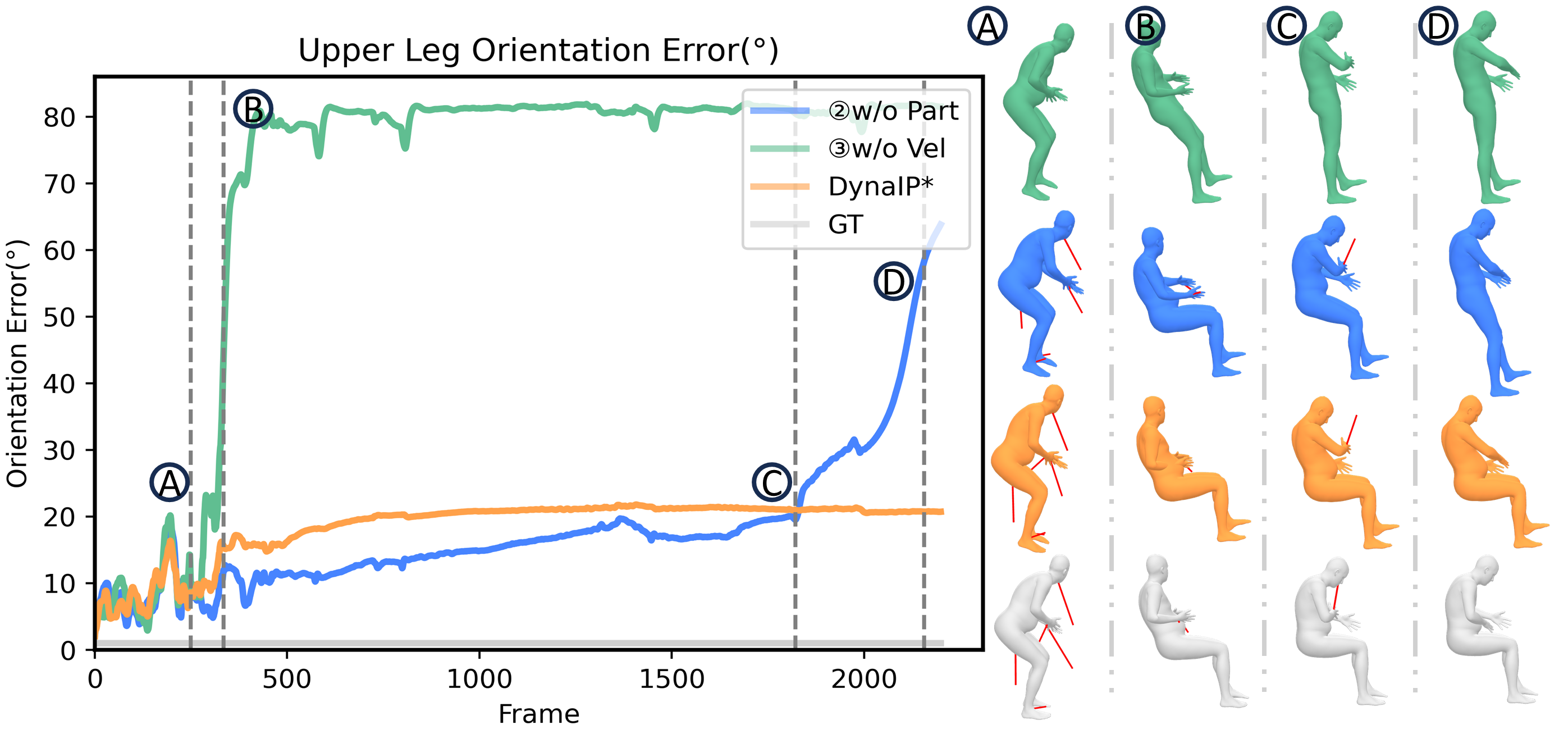}
	\caption{Visualization of upper leg orientation error over time on a test sequence from Natural Motion~\cite{geissinger2020motion}.}
	\label{fig:leg_error_over_time}
	\vspace{-5mm}
\end{figure}

To evaluate the effectiveness of key components in our model, we compared three additional variants with our method:
\ding{172}Baseline: a naive LSTM network with RNN-initialization~\cite{yi2022physical}; 
\ding{173}w/o Part: same as DynaIP* but without partition modeling; \ding{174}w/o Vel: same as DynaIP* but without learning velocity. 
The results of these variants on CIP, DIP-IMU, and Natural Motion in Tab.~\ref{table:ablation} demonstrate a clear increase in testing errors as each component is progressively removed. This trend emphasizes the significance of both pseudo velocity learning and part-based modeling for achieving robust human pose estimation. These components work in synergy to improve the overall performance of our model.
The visualization of a test sequence from Natural Motion in Fig.~\ref{fig:leg_error_over_time} highlights the performance differences between the variants. Variant\ding{174} initially struggles to perform a sitting-down motion, while variant\ding{173}, which includes velocity modeling, can perform the sitting motion but experiences gradual drift over time due to sensor noise. In contrast, our complete model (DynaIP*) combines both velocity and part-based modeling, resulting in accurate and robust pose estimation throughout the sequence. This demonstrates the effectiveness of our approach in handling complex and dynamic motions. 

\section{Conclusions}
This study focuses on improving the robustness and accuracy of learning-based HPE utilizing sparse inertial sensors. The methodology involves the integration of real-world motion capture data from diverse skeleton formats. It employs pseudo-velocity as an intermediary and introduces a novel part-based approach. This method effectively leverages acceleration data and local body correlations, leading to enhanced pose estimation results. As evidenced in the case studies, this method significantly outperforms existing techniques on all performance metrics across five datasets.

\clearpage
{
    \small
    \bibliographystyle{ieeenat_fullname}
    \bibliography{main}

\begin{thebibliography}{51}
\providecommand{\natexlab}[1]{#1}
\providecommand{\url}[1]{\texttt{#1}}
\expandafter\ifx\csname urlstyle\endcsname\relax
  \providecommand{\doi}[1]{doi: #1}\else
  \providecommand{\doi}{doi: \begingroup \urlstyle{rm}\Url}\fi

\bibitem[Aberman et~al.(2020)Aberman, Li, Lischinski, Sorkine-Hornung,
  Cohen-Or, and Chen]{aberman2020skeleton}
Kfir Aberman, Peizhuo Li, Dani Lischinski, Olga Sorkine-Hornung, Daniel
  Cohen-Or, and Baoquan Chen.
\newblock Skeleton-aware networks for deep motion retargeting.
\newblock \emph{ACM Transactions on Graphics (TOG)}, 39\penalty0 (4):\penalty0
  62--1, 2020.

\bibitem[An et~al.(2022)An, Li, and Ogras]{an2022mri}
Sizhe An, Yin Li, and Umit Ogras.
\newblock mri: Multi-modal 3d human pose estimation dataset using mmwave,
  rgb-d, and inertial sensors.
\newblock \emph{Advances in Neural Information Processing Systems(NeurIPS)},
  35:\penalty0 27414--27426, 2022.

\bibitem[Chen et~al.(2022)Chen, Wang, Zhu, Li, Chen, and Ye]{chen2022mmbody}
Anjun Chen, Xiangyu Wang, Shaohao Zhu, Yanxu Li, Jiming Chen, and Qi Ye.
\newblock mmbody benchmark: 3d body reconstruction dataset and analysis for
  millimeter wave radar.
\newblock In \emph{Proceedings of the 30th ACM International Conference on
  Multimedia (ACM MM)}, pages 3501--3510, 2022.

\bibitem[Christensen et~al.(2023)Christensen, Fernandez, Smith, Michalareas,
  Yazdi, Farahi, Schmidt, Bahmanian, and Roig]{emokine}
Julia~F. Christensen, Andres Fernandez, Rebecca Smith, Georgios Michalareas,
  Sina H.~N. Yazdi, Fahima Farahi, Eva-Madeleine Schmidt, Nasimeh Bahmanian,
  and Gemma Roig.
\newblock Emokine: A kinematic dataset and computational framework for scaling
  up the creation of highly controlled emotional full-body movement datasets.
\newblock \url{https://doi.org/10.5281/zenodo.7821844}, 2023.

\bibitem[Dai et~al.(2022)Dai, Lin, Wen, Shen, Xu, Yu, Ma, and
  Wang]{dai2022hsc4d}
Yudi Dai, Yitai Lin, Chenglu Wen, Siqi Shen, Lan Xu, Jingyi Yu, Yuexin Ma, and
  Cheng Wang.
\newblock Hsc4d: Human-centered 4d scene capture in large-scale indoor-outdoor
  space using wearable imus and lidar.
\newblock In \emph{Proceedings of the IEEE/CVF Conference on Computer Vision
  and Pattern Recognition(CVPR)}, pages 6792--6802, 2022.

\bibitem[Dai et~al.(2023)Dai, Lin, Lin, Wen, Xu, Yi, Shen, Ma, and
  Wang]{dai2023sloper4d}
Yudi Dai, YiTai Lin, XiPing Lin, Chenglu Wen, Lan Xu, Hongwei Yi, Siqi Shen,
  Yuexin Ma, and Cheng Wang.
\newblock Sloper4d: A scene-aware dataset for global 4d human pose estimation
  in urban environments.
\newblock In \emph{Proceedings of the IEEE/CVF Conference on Computer Vision
  and Pattern Recognition(CVPR)}, pages 682--692, 2023.

\bibitem[Geissinger and Asbeck(2020)]{geissinger2020motion}
Jack~H Geissinger and Alan~T Asbeck.
\newblock Motion inference using sparse inertial sensors, self-supervised
  learning, and a new dataset of unscripted human motion.
\newblock \url{https://doi.org/10.7294/2v3w-sb92}, 2020.

\bibitem[Geng et~al.(2021)Geng, Sun, Xiao, Zhang, and Wang]{geng2021bottom}
Zigang Geng, Ke Sun, Bin Xiao, Zhaoxiang Zhang, and Jingdong Wang.
\newblock Bottom-up human pose estimation via disentangled keypoint regression.
\newblock In \emph{Proceedings of the IEEE/CVF conference on computer vision
  and pattern recognition(CVPR)}, pages 14676--14686, 2021.

\bibitem[Guidolin et~al.(2022)Guidolin, Menegatti, and
  Reggiani]{guidolin2022unipd}
Mattia Guidolin, Emanuele Menegatti, and Monica Reggiani.
\newblock Unipd-bpe: Synchronized rgb-d and inertial data for multimodal body
  pose estimation and tracking.
\newblock \url{https://doi.org/10.17605/OSF.IO/YJ9Q4}, 2022.

\bibitem[Guzov et~al.(2021)Guzov, Mir, Sattler, and Pons-Moll]{guzov2021human}
Vladimir Guzov, Aymen Mir, Torsten Sattler, and Gerard Pons-Moll.
\newblock Human poseitioning system (hps): 3d human pose estimation and
  self-localization in large scenes from body-mounted sensors.
\newblock In \emph{Proceedings of the IEEE/CVF Conference on Computer Vision
  and Pattern Recognition(CVPR)}, pages 4318--4329, 2021.

\bibitem[Guzov et~al.(2022)Guzov, Sattler, and Pons-Moll]{guzov2022visually}
Vladimir Guzov, Torsten Sattler, and Gerard Pons-Moll.
\newblock Visually plausible human-object interaction capture from wearable
  sensors.
\newblock \emph{arXiv preprint arXiv:2205.02830}, 2022.

\bibitem[Herath et~al.(2020)Herath, Yan, and Furukawa]{herath2020ronin}
Sachini Herath, Hang Yan, and Yasutaka Furukawa.
\newblock Ronin: Robust neural inertial navigation in the wild: Benchmark,
  evaluations, \& new methods.
\newblock In \emph{2020 IEEE International Conference on Robotics and
  Automation (ICRA)}, pages 3146--3152. IEEE, 2020.

\bibitem[Hoyer et~al.(2022)Hoyer, Dai, and Van~Gool]{hoyer2022daformer}
Lukas Hoyer, Dengxin Dai, and Luc Van~Gool.
\newblock Daformer: Improving network architectures and training strategies for
  domain-adaptive semantic segmentation.
\newblock In \emph{Proceedings of the IEEE/CVF Conference on Computer Vision
  and Pattern Recognition(CVPR)}, pages 9924--9935, 2022.

\bibitem[Huang et~al.(2018)Huang, Kaufmann, Aksan, Black, Hilliges, and
  Pons-Moll]{huang2018deep}
Yinghao Huang, Manuel Kaufmann, Emre Aksan, Michael~J Black, Otmar Hilliges,
  and Gerard Pons-Moll.
\newblock Deep inertial poser: Learning to reconstruct human pose from sparse
  inertial measurements in real time.
\newblock \emph{ACM Transactions on Graphics (TOG)}, 37\penalty0 (6):\penalty0
  1--15, 2018.

\bibitem[Jiang et~al.(2022{\natexlab{a}})Jiang, Streli, Qiu, Fender, Laich,
  Snape, and Holz]{jiang2022avatarposer}
Jiaxi Jiang, Paul Streli, Huajian Qiu, Andreas Fender, Larissa Laich, Patrick
  Snape, and Christian Holz.
\newblock Avatarposer: Articulated full-body pose tracking from sparse motion
  sensing.
\newblock In \emph{Proceedings of European Conference on Computer
  Vision(ECCV)}. Springer, 2022{\natexlab{a}}.

\bibitem[Jiang et~al.(2022{\natexlab{b}})Jiang, Ye, Gopinath, Won, Winkler, and
  Liu]{jiang2022transformer}
Yifeng Jiang, Yuting Ye, Deepak Gopinath, Jungdam Won, Alexander~W Winkler, and
  C~Karen Liu.
\newblock Transformer inertial poser: Real-time human motion reconstruction
  from sparse imus with simultaneous terrain generation.
\newblock In \emph{SIGGRAPH Asia 2022 Conference Papers(SA' 22)}, pages 1--9,
  2022{\natexlab{b}}.

\bibitem[Kaufmann et~al.(2021)Kaufmann, Zhao, Tang, Tao, Twigg, Song, Wang, and
  Hilliges]{kaufmann2021pose}
Manuel Kaufmann, Yi Zhao, Chengcheng Tang, Lingling Tao, Christopher Twigg, Jie
  Song, Robert Wang, and Otmar Hilliges.
\newblock Em-pose: 3d human pose estimation from sparse electromagnetic
  trackers.
\newblock In \emph{Proceedings of the IEEE/CVF international conference on
  computer vision(ICCV)}, pages 11510--11520, 2021.

\bibitem[Kocabas et~al.(2020)Kocabas, Athanasiou, and Black]{kocabas2020vibe}
Muhammed Kocabas, Nikos Athanasiou, and Michael~J Black.
\newblock Vibe: Video inference for human body pose and shape estimation.
\newblock In \emph{Proceedings of the IEEE/CVF conference on computer vision
  and pattern recognition(CVPR)}, pages 5253--5263, 2020.

\bibitem[Kwon et~al.(2020)Kwon, Tong, Haresamudram, Gao, Abowd, Lane, and
  Ploetz]{kwon2020imutube}
Hyeokhyen Kwon, Catherine Tong, Harish Haresamudram, Yan Gao, Gregory~D Abowd,
  Nicholas~D Lane, and Thomas Ploetz.
\newblock Imutube: Automatic extraction of virtual on-body accelerometry from
  video for human activity recognition.
\newblock \emph{Proceedings of the ACM on Interactive, Mobile, Wearable and
  Ubiquitous Technologies(IMWUT)}, 4\penalty0 (3):\penalty0 1--29, 2020.

\bibitem[Lee and Lee(2021)]{lee2021uncertainty}
Gun-Hee Lee and Seong-Whan Lee.
\newblock Uncertainty-aware human mesh recovery from video by learning
  part-based 3d dynamics.
\newblock In \emph{Proceedings of the IEEE/CVF International Conference on
  Computer Vision(ICCV)}, pages 12375--12384, 2021.

\bibitem[Loper et~al.(2014)Loper, Mahmood, and Black]{loper2014mosh}
Matthew Loper, Naureen Mahmood, and Michael~J Black.
\newblock Mosh: motion and shape capture from sparse markers.
\newblock \emph{ACM Trans. Graph. (TOG)}, 33\penalty0 (6):\penalty0 220--1,
  2014.

\bibitem[Loper et~al.(2023)Loper, Mahmood, Romero, Pons-Moll, and
  Black]{loper2023smpl}
Matthew Loper, Naureen Mahmood, Javier Romero, Gerard Pons-Moll, and Michael~J
  Black.
\newblock Smpl: A skinned multi-person linear model.
\newblock In \emph{Seminal Graphics Papers: Pushing the Boundaries, Volume 2},
  pages 851--866. 2023.

\bibitem[Ma et~al.(2023)Ma, Su, Wang, Zhu, and Wang]{ma20233d}
Xiaoxuan Ma, Jiajun Su, Chunyu Wang, Wentao Zhu, and Yizhou Wang.
\newblock 3d human mesh estimation from virtual markers.
\newblock In \emph{Proceedings of the IEEE/CVF Conference on Computer Vision
  and Pattern Recognition (CVPR)}, pages 534--543, 2023.

\bibitem[Mahmood et~al.(2019)Mahmood, Ghorbani, Troje, Pons-Moll, and
  Black]{mahmood2019amass}
Naureen Mahmood, Nima Ghorbani, Nikolaus~F Troje, Gerard Pons-Moll, and
  Michael~J Black.
\newblock Amass: Archive of motion capture as surface shapes.
\newblock In \emph{Proceedings of the IEEE/CVF international conference on
  computer vision(ICCV)}, pages 5442--5451, 2019.

\bibitem[Maurice et~al.(2019)Maurice, Malais{\'e}, Amiot, Paris, Richard,
  Rochel, and Ivaldi]{maurice2019human}
Pauline Maurice, Adrien Malais{\'e}, Cl{\'e}lie Amiot, Nicolas Paris,
  Guy-Junior Richard, Olivier Rochel, and Serena Ivaldi.
\newblock Human movement and ergonomics: An industry-oriented dataset for
  collaborative robotics.
\newblock \emph{The International Journal of Robotics Research(IJRR)},
  38\penalty0 (14):\penalty0 1529--1537, 2019.

\bibitem[Mollyn et~al.(2023)Mollyn, Arakawa, Goel, Harrison, and
  Ahuja]{mollyn2023imuposer}
Vimal Mollyn, Riku Arakawa, Mayank Goel, Chris Harrison, and Karan Ahuja.
\newblock Imuposer: Full-body pose estimation using imus in phones, watches,
  and earbuds.
\newblock In \emph{Proceedings of the 2023 CHI Conference on Human Factors in
  Computing Systems(CHI)}, pages 1--12, 2023.

\bibitem[Mourot et~al.(2023)Mourot, Hoyet, Clerc, and Hellier]{mourot2023humor}
Lucas Mourot, Ludovic Hoyet, Fran{\c{c}}ois~Le Clerc, and Pierre Hellier.
\newblock Humor: Human motion representation using topology-agnostic
  transformers for character animation retargeting.
\newblock \emph{arXiv preprint arXiv:2305.18897}, 2023.

\bibitem[Murray et~al.(2017)Murray, Li, and Sastry]{murray2017mathematical}
Richard~M Murray, Zexiang Li, and S~Shankar Sastry.
\newblock \emph{A mathematical introduction to robotic manipulation}.
\newblock CRC press, 2017.

\bibitem[Palermo et~al.(2022)Palermo, Cerqueira, Andr{\'e}, Pereira, and
  Santos]{palermo2022raw}
Manuel Palermo, Sara~M Cerqueira, Jo{\~a}o Andr{\'e}, Ant{\'o}nio Pereira, and
  Cristina~P Santos.
\newblock From raw measurements to human pose-a dataset with low-cost and
  high-end inertial-magnetic sensor data.
\newblock \emph{Scientific Data}, 9\penalty0 (1):\penalty0 591, 2022.

\bibitem[Pei et~al.(2021)Pei, Xia, Chu, Xiao, Wu, Yu, and Qiu]{pei2021mars}
Ling Pei, Songpengcheng Xia, Lei Chu, Fanyi Xiao, Qi Wu, Wenxian Yu, and Robert
  Qiu.
\newblock Mars: Mixed virtual and real wearable sensors for human activity
  recognition with multidomain deep learning model.
\newblock \emph{IEEE Internet of Things Journal}, 8\penalty0 (11):\penalty0
  9383--9396, 2021.

\bibitem[Peng et~al.(2023)Peng, Zheng, and Chen]{peng2023source}
Qucheng Peng, Ce Zheng, and Chen Chen.
\newblock Source-free domain adaptive human pose estimation.
\newblock In \emph{Proceedings of the IEEE/CVF International Conference on
  Computer Vision (ICCV)}, pages 4826--4836, 2023.

\bibitem[Ren et~al.(2023)Ren, Zhao, He, Cong, Liang, Yu, Xu, and
  Ma]{ren2023lidar}
Yiming Ren, Chengfeng Zhao, Yannan He, Peishan Cong, Han Liang, Jingyi Yu, Lan
  Xu, and Yuexin Ma.
\newblock Lidar-aid inertial poser: Large-scale human motion capture by sparse
  inertial and lidar sensors.
\newblock \emph{IEEE Transactions on Visualization and Computer Graphics},
  29\penalty0 (5):\penalty0 2337--2347, 2023.

\bibitem[S{\'a}r{\'a}ndi et~al.(2023)S{\'a}r{\'a}ndi, Hermans, and
  Leibe]{sarandi2023learning}
Istv{\'a}n S{\'a}r{\'a}ndi, Alexander Hermans, and Bastian Leibe.
\newblock Learning 3d human pose estimation from dozens of datasets using a
  geometry-aware autoencoder to bridge between skeleton formats.
\newblock In \emph{Proceedings of the IEEE/CVF Winter Conference on
  Applications of Computer Vision(WACV)}, pages 2956--2966, 2023.

\bibitem[Schepers et~al.(2018)Schepers, Giuberti, Bellusci,
  et~al.]{schepers2018xsens}
Martin Schepers, Matteo Giuberti, Giovanni Bellusci, et~al.
\newblock Xsens mvn: Consistent tracking of human motion using inertial
  sensing.
\newblock \emph{Xsens Technol}, 1\penalty0 (8):\penalty0 1--8, 2018.

\bibitem[Shen et~al.(2023)Shen, Cen, Peng, Shuai, Bao, and
  Zhou]{shen2023learning}
Zehong Shen, Zhi Cen, Sida Peng, Qing Shuai, Hujun Bao, and Xiaowei Zhou.
\newblock Learning human mesh recovery in 3d scenes.
\newblock In \emph{Proceedings of the IEEE/CVF Conference on Computer Vision
  and Pattern Recognition(CVPR)}, pages 17038--17047, 2023.

\bibitem[Shuai et~al.(2022)Shuai, Wu, and Liu]{shuai2022adaptive}
Hui Shuai, Lele Wu, and Qingshan Liu.
\newblock Adaptive multi-view and temporal fusing transformer for 3d human pose
  estimation.
\newblock \emph{IEEE Transactions on Pattern Analysis and Machine
  Intelligence(TPAMI)}, 45\penalty0 (4):\penalty0 4122--4135, 2022.

\bibitem[Slyper and Hodgins(2008)]{slyper2008action}
Ronit Slyper and Jessica~K Hodgins.
\newblock Action capture with accelerometers.
\newblock In \emph{Proceedings of the 2008 ACM SIGGRAPH/Eurographics symposium
  on computer animation}, pages 193--199, 2008.

\bibitem[Tautges et~al.(2011)Tautges, Zinke, Krüger, Baumann, Weber, Helten,
  Müller, Seidel, and Eberhardt]{Tautges_2011}
Jochen Tautges, Arno Zinke, Björn Krüger, Jan Baumann, Andreas Weber, Thomas
  Helten, Meinard Müller, Hans-Peter Seidel, and Bernd Eberhardt.
\newblock Motion reconstruction using sparse accelerometer data.
\newblock \emph{ACM Transactions on Graphics(TOG)}, 30\penalty0 (3):\penalty0
  1–12, 2011.

\bibitem[Tripathi et~al.(2023)Tripathi, M{\"u}ller, Huang, Taheri, Black, and
  Tzionas]{tripathi20233d}
Shashank Tripathi, Lea M{\"u}ller, Chun-Hao~P Huang, Omid Taheri, Michael~J
  Black, and Dimitrios Tzionas.
\newblock 3d human pose estimation via intuitive physics.
\newblock In \emph{Proceedings of the IEEE/CVF Conference on Computer Vision
  and Pattern Recognition(CVPR)}, pages 4713--4725, 2023.

\bibitem[Trumble et~al.(2017)Trumble, Gilbert, Malleson, Hilton, and
  Collomosse]{trumble2017total}
Matthew Trumble, Andrew Gilbert, Charles Malleson, Adrian Hilton, and John
  Collomosse.
\newblock Total capture: 3d human pose estimation fusing video and inertial
  sensors.
\newblock In \emph{Proceedings of 28th British Machine Vision
  Conference(BMVC)}, pages 1--13, 2017.

\bibitem[Von~Marcard et~al.(2017)Von~Marcard, Rosenhahn, Black, and
  Pons-Moll]{von2017sparse}
Timo Von~Marcard, Bodo Rosenhahn, Michael~J Black, and Gerard Pons-Moll.
\newblock Sparse inertial poser: Automatic 3d human pose estimation from sparse
  imus.
\newblock In \emph{Computer graphics forum(CGF)}, pages 349--360. Wiley Online
  Library, 2017.

\bibitem[Xia et~al.(2021)Xia, Chu, Pei, Zhang, Yu, and Qiu]{xia2021learning}
Songpengcheng Xia, Lei Chu, Ling Pei, Zixuan Zhang, Wenxian Yu, and Robert~C
  Qiu.
\newblock Learning disentangled representation for mixed-reality human activity
  recognition with a single imu sensor.
\newblock \emph{IEEE Transactions on Instrumentation and Measurement},
  70:\penalty0 1--14, 2021.

\bibitem[Xue et~al.(2022)Xue, Cao, Ju, Hu, Wang, Zhang, and Su]{xue2022m4esh}
Hongfei Xue, Qiming Cao, Yan Ju, Haochen Hu, Haoyu Wang, Aidong Zhang, and Lu
  Su.
\newblock M4esh: mmwave-based 3d human mesh construction for multiple subjects.
\newblock In \emph{Proceedings of the 20th ACM Conference on Embedded Networked
  Sensor Systems(SenSys)}, pages 391--406, 2022.

\bibitem[Yang et~al.(2021)Yang, Kim, and Lee]{yang2021lobstr}
Dongseok Yang, Doyeon Kim, and Sung-Hee Lee.
\newblock Lobstr: Real-time lower-body pose prediction from sparse upper-body
  tracking signals.
\newblock In \emph{Computer Graphics Forum(CGF)}, pages 265--275. Wiley Online
  Library, 2021.

\bibitem[Yi et~al.(2021)Yi, Zhou, and Xu]{yi2021transpose}
Xinyu Yi, Yuxiao Zhou, and Feng Xu.
\newblock Transpose: Real-time 3d human translation and pose estimation with
  six inertial sensors.
\newblock \emph{ACM Transactions on Graphics (TOG)}, 40\penalty0 (4):\penalty0
  1--13, 2021.

\bibitem[Yi et~al.(2022)Yi, Zhou, Habermann, Shimada, Golyanik, Theobalt, and
  Xu]{yi2022physical}
Xinyu Yi, Yuxiao Zhou, Marc Habermann, Soshi Shimada, Vladislav Golyanik,
  Christian Theobalt, and Feng Xu.
\newblock Physical inertial poser (pip): Physics-aware real-time human motion
  tracking from sparse inertial sensors.
\newblock In \emph{Proceedings of the IEEE/CVF Conference on Computer Vision
  and Pattern Recognition(CVPR)}, pages 13167--13178, 2022.

\bibitem[Zeng et~al.(2020)Zeng, Sun, Huang, Liu, Xu, and Lin]{zeng2020srnet}
Ailing Zeng, Xiao Sun, Fuyang Huang, Minhao Liu, Qiang Xu, and Stephen Lin.
\newblock Srnet: Improving generalization in 3d human pose estimation with a
  split-and-recombine approach.
\newblock In \emph{Proceedings of European Conference on Computer
  Vision(ECCV)}, pages 507--523. Springer, 2020.

\bibitem[Zhang and Gao(2022)]{zhang2022transfer}
Lei Zhang and Xinbo Gao.
\newblock Transfer adaptation learning: A decade survey.
\newblock \emph{IEEE Transactions on Neural Networks and Learning Systems},
  2022.

\bibitem[Zhang et~al.(2020)Zhang, Wang, Qin, and Zeng]{zhang2020fusing}
Zhe Zhang, Chunyu Wang, Wenhu Qin, and Wenjun Zeng.
\newblock Fusing wearable imus with multi-view images for human pose
  estimation: A geometric approach.
\newblock In \emph{Proceedings of the IEEE/CVF Conference on Computer Vision
  and Pattern Recognition(CVPR)}, pages 2200--2209, 2020.

\bibitem[Zhao et~al.(2018)Zhao, Li, Abu~Alsheikh, Tian, Zhao, Torralba, and
  Katabi]{zhao2018through}
Mingmin Zhao, Tianhong Li, Mohammad Abu~Alsheikh, Yonglong Tian, Hang Zhao,
  Antonio Torralba, and Dina Katabi.
\newblock Through-wall human pose estimation using radio signals.
\newblock In \emph{Proceedings of the IEEE conference on computer vision and
  pattern recognition(CVPR)}, pages 7356--7365, 2018.

\bibitem[Zheng et~al.(2023)Zheng, Wu, Chen, Yang, Zhu, Shen, Kehtarnavaz, and
  Shah]{zheng2023deep}
Ce Zheng, Wenhan Wu, Chen Chen, Taojiannan Yang, Sijie Zhu, Ju Shen, Nasser
  Kehtarnavaz, and Mubarak Shah.
\newblock Deep learning-based human pose estimation: A survey.
\newblock \emph{ACM Computing Surveys}, 56\penalty0 (1):\penalty0 1--37, 2023.

\end{thebibliography}
}


\clearpage
\setcounter{page}{1}

\twocolumn[
	\begin{@twocolumnfalse}
		\section*{\centering{\Large{Dynamic Inertial Poser (DynaIP): Part-Based Motion Dynamics Learning for Enhanced Human Pose Estimation with Sparse Inertial Sensors}}}
		\begin{center}
			\large{Supplementary Material}


		\end{center}
		\centering
	\end{@twocolumnfalse}
]

\setcounter{section}{0}
\renewcommand\thesection{\Alph{section}}


This supplementary material provides additional information to complement our main paper. Sec.~\ref{sec:implementation} outlines the network architecture and training details of our approach. Sec.~\ref{sec:datasets} offers an overview of the datasets utilized in our research. Sec.~\ref{sec:add_analysis} presents extended ablation studies, encompassing various training data settings and further analysis of component design. Sec.~\ref{sec:add_qualitative} features additional qualitative comparisons between our model and existing state-of-the-art methods. Finally, Sec.~\ref{sec:discussion} summarizes our work, discusses its limitations, and proposes potential directions for future research.

\section{Implementation Details}
\label{sec:implementation}
\textbf{Network Details} Our model architecture comprises distinct sub-networks, each equipped with a Multilayer Perceptron (MLP) layer for initializing hidden states, followed by the Long Short-Term Memory (LSTM) networks. The first two-layer LSTM network is designed for regressing the pseudo-velocity of joints, and the second two-layer LSTM network focuses on the final pose estimation. An additional LSTM layer is used to extract global context from all six sensors, which is shared by all three sub-networks. Notably, the output for each joint is represented as a global 6D-rotation relative to the root joint~\cite{yi2021transpose}, and this relational approach is also applied to intermediate pseudo-velocity predictions.

Specifically, we directly regress the orientations of 11 joints on the Xsens skeleton that lack IMU measurements. To align with the SMPL model during evaluation (on DIP-IMU) and visualization, we remove one redundant torso joint (labeled as 'L3' in our implementation) and map the remaining predicted results. When integrating with the DIP-IMU dataset, we duplicate the 'Spine1' joint in the SMPL model to correspond with our model's output dimensions.

A difference from prior methodologies in our work is the treatment of joints such as the head, root, forearms, and forelegs, which are directly attached to IMUs. Rather than predicting their orientations, we directly utilize the orientations provided by their respective IMU measurements, ensuring a more straightforward and accurate representation.



\begin{table}[!t]
\resizebox{0.46\textwidth}{!}{%
\begin{tabular}{@{}llllllll@{}}
\toprule
Datasets & Motion types & Subjects & Minutes\\
\midrule
DIP-IMU~\cite{huang2018deep} & jumping, sitting, walking, lifting arm & 10 & 86\\
CIP~\cite{palermo2022raw} & grabbing, reaching, sitting & 10 & 288\\
Natural Motion~\cite{geissinger2020motion} & long time sitting, exercising, walking & 17 & 692\\
Emokine~\cite{emokine} & upper body motion & 1 & 12\\
AnDy~\cite{maurice2019human} & walking, kneeling, manipulating loads  & 13 & 421\\
UNIPD~\cite{guidolin2022unipd} & sitting, pointing, bending, walking and jogging & 15 & 162\\
\midrule
Real IMU data & - & - & 1661\\
AMASS~\cite{mahmood2019amass} & - & - & 2122 \\
\bottomrule
\end{tabular}
}
\caption{Dataset Overview}
\label{table:datasets_overview}
\vspace{-5mm}
\end{table}
\section{Datasets Details}
\label{sec:datasets}

\begin{table*}[!t]
\centering
\resizebox{1.0\textwidth}{!}{
\begin{tabular}{l*{10}c}
\toprule
\multirow{2}{*}{} & \multicolumn{2}{c}{DIP IMU} & \multicolumn{2}{c}{AnDy} & \multicolumn{2}{c}{UNIPD} & \multicolumn{2}{c}{CIP} & \multicolumn{2}{c}{Natural Motion} \\
\cmidrule(lr){2-3} \cmidrule(lr){4-5} \cmidrule(lr){6-7} \cmidrule(lr){8-9} \cmidrule(lr){10-11}
 & SIP Err(°) & Ang Err(°) & SIP Err(°) & Ang Err(°) & SIP Err(°) & Ang Err(°) & SIP Err(°) & Ang Err(°) & SIP Err(°) & Ang Err(°) \\
\midrule
\textit{1)}AMASS & 23.80 & 8.25 & 20.80& 7.72& 14.98& 5.64& 25.27& 9.08& 40.58& 13.56\\
\textit{2)}AMASS+DIP & 14.41& \underline{5.90}& 18.79& 8.10& 14.29& 5.49& 18.65& 7.51& 20.72& 9.26\\
\textit{3)}Xsens & 17.62& 8.33& \underline{8.93}& \underline{3.45}& \underline{7.29}& \underline{2.77}& \textbf{11.42}& \textbf{4.54}& \textbf{15.78}& \textbf{7.18}\\
\textit{4)}Xsens+DIP & \textbf{13.67}& \textbf{5.83}& 9.17& 3.56 & 7.60& 2.83& \underline{11.67}& \underline{4.63}& \underline{18.88}& \underline{8.03}\\

\textit{5)}AMASS+Xsens+DIP & \underline{13.77}& 5.92 & \textbf{8.84}& \textbf{3.42} & \textbf{7.08}& \textbf{2.67}& 12.01& 4.66& 19.89& 8.39\\

\bottomrule
\end{tabular}
}
\caption{Performance Comparison on Different Training Data Settings.}
\label{table:more_v_to_r}
\end{table*}

\begin{table*}[!t]
\centering
\resizebox{1.0\textwidth}{!}{
\begin{tabular}{l*{10}c}
\toprule
\multirow{2}{*}{} & \multicolumn{2}{c}{DIP IMU} & \multicolumn{2}{c}{AnDy} & \multicolumn{2}{c}{UNIPD} & \multicolumn{2}{c}{CIP} & \multicolumn{2}{c}{Natural Motion} \\
\cmidrule(lr){2-3} \cmidrule(lr){4-5} \cmidrule(lr){6-7} \cmidrule(lr){8-9} \cmidrule(lr){10-11}
 & SIP Err(°) & Ang Err(°) & SIP Err(°) & Ang Err(°) & SIP Err(°) & Ang Err(°) & SIP Err(°) & Ang Err(°) & SIP Err(°) & Ang Err(°) \\
\midrule
\ding{172}Baseline & 15.26 & 6.17 & 9.89& 3.76& 9.05& 3.24& 13.18& 5.13& 33.22& 11.20\\
\ding{173}w/o Part & 14.97& \underline{6.02}& 9.62& 3.85& 7.63& 2.92& 13.00& \underline{4.86}& 31.62& 9.60\\
\ding{174}w/o Vel & 14.87& 6.11& \underline{9.27}& \textbf{3.50}& 7.61& \textbf{2.81}& 12.54& 4.89& 29.15& 10.49\\
\ding{175}CP & \underline{14.64}& 6.35& 10.74& 4.14& \textbf{7.54}& 2.88& \underline{12.52}& 4.87& \underline{20.58}& \textbf{7.90}\\
\ding{176}DynaIP* & \textbf{13.67}& \textbf{5.83}& \textbf{9.17}& \underline{3.56}& \underline{7.60}& \underline{2.83}& \textbf{11.67}& \textbf{4.63}& \textbf{18.88}& \underline{8.03}\\
\bottomrule
\end{tabular}
}
\caption{Performance Comparison of Ablation Variants.}
\label{table:more_ablation_comparison}
\end{table*}

We use a subset of AMASS~\cite{mahmood2019amass} to generate synthetic IMU as previous works. Simultaneously, we train and evaluate our model using several publicly available Xsens Inertial Mocap dataset. The following part will briefly introduce these public datasets we used.

\textit{1)} AnDy~\cite{maurice2019human}: The Inertial Mocap data from AnDy were collected from 13 participants in an industry environment. A total number of 195 trails consist of commonly seen motions like raising arms, walking, bending torso, crouching and etc.. We use the last two subject's data (ID: 9266 and 9857) for evaluation and the others for training.

\textit{2)} CIP~\cite{palermo2022raw}: The sensing data of the CIP dataset were collected from 10 participants for 6 different types of movement sequences. These sequence types involve motions such as factory assembly tasks, office dynamics and random free movements. We select data from the subject 4 and 8 for evaluation and the rest data for training.

\textit{3)} Natural Motion~\cite{geissinger2020motion}: The dataset consists of motions from 17 participants, with 13 performing unscripted daily activities like walking, working at a computer, exercising, etc., and the remaining 4 executing actions within industrial settings. Notably, an essential characteristic of this dataset is the exceptional duration of each raw capture sequence, varying from half an hour up to three hours. This allows for a wealth of extended long-time sitting or standing sequences to be incorporated. Nevertheless, we find some raw data may exhibit certain drift due to prolonged capture times and lack of precise re-calibration. We manually selected 9 (ID: 1, 2, 3, 4, 5, 6, 10, 11 and 13) out of the 13 participants engaged in daily activities and extracted clean data segments to be utilized for training and testing.

\textit{4)} Emokine~\cite{emokine}: This dataset contains 63 sequences captured from a dancer performing different body movements with emotions like anger, contentment, fear, joy, neutrality, and sadness, this includes a variety of rapid and slow movements of the upper and lower limbs. Since the total duration of this data only amounts to 12 minutes, we have decided to use it solely for training purposes.

\textit{5)} UNIPD~\cite{guidolin2022unipd}: This dataset captures detailed body poses from 15 participants performing 12 scripted activities in laborious environment, such as walking, sitting, jogging and bending.  We use data from last two subjects (ID: 14 and 15) as test set and the rest for training.

The overview of these dataset along with DIP-IMU~\cite{huang2018deep} is shown in Tab.~\ref{table:datasets_overview}

\begin{figure}[!t]
	\includegraphics[width=8cm]{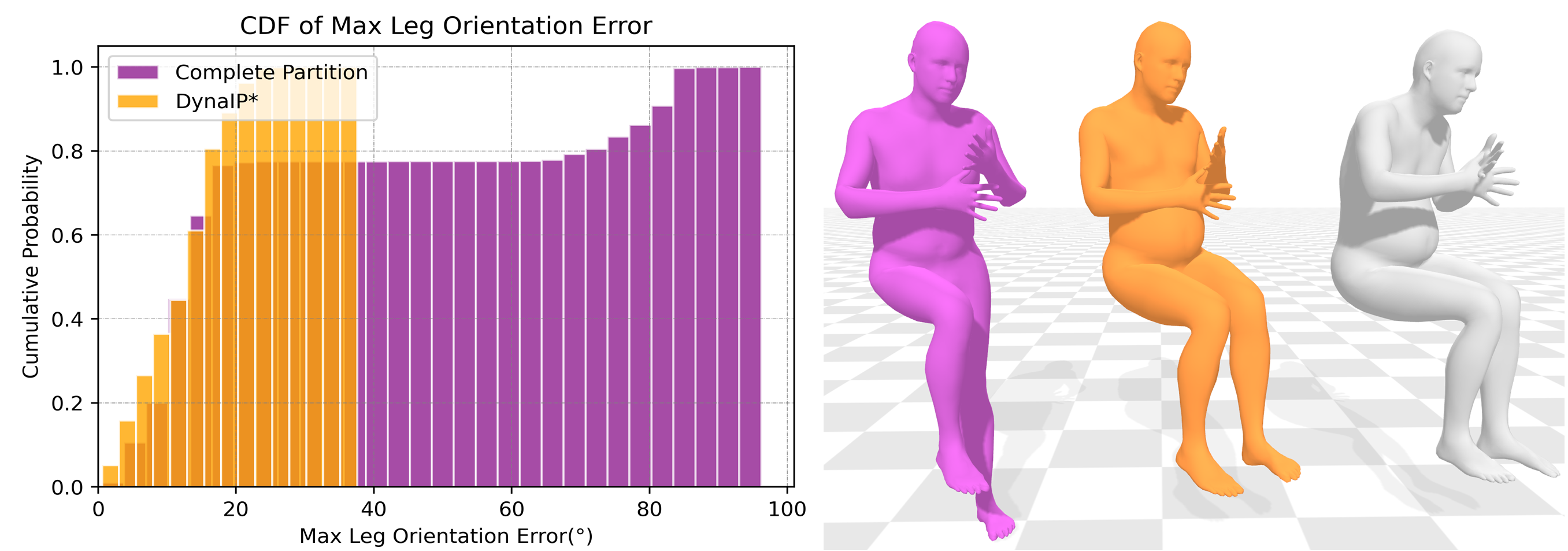}
	\caption{CDF of the largest upper leg orientation error. Complete partition model predicts unnatural result for excessive separation.}
	\label{fig:partition_ablation}
	\vspace{-5mm}
\end{figure}

\begin{figure*}[!t]
    \centering
	\includegraphics[width=16cm]{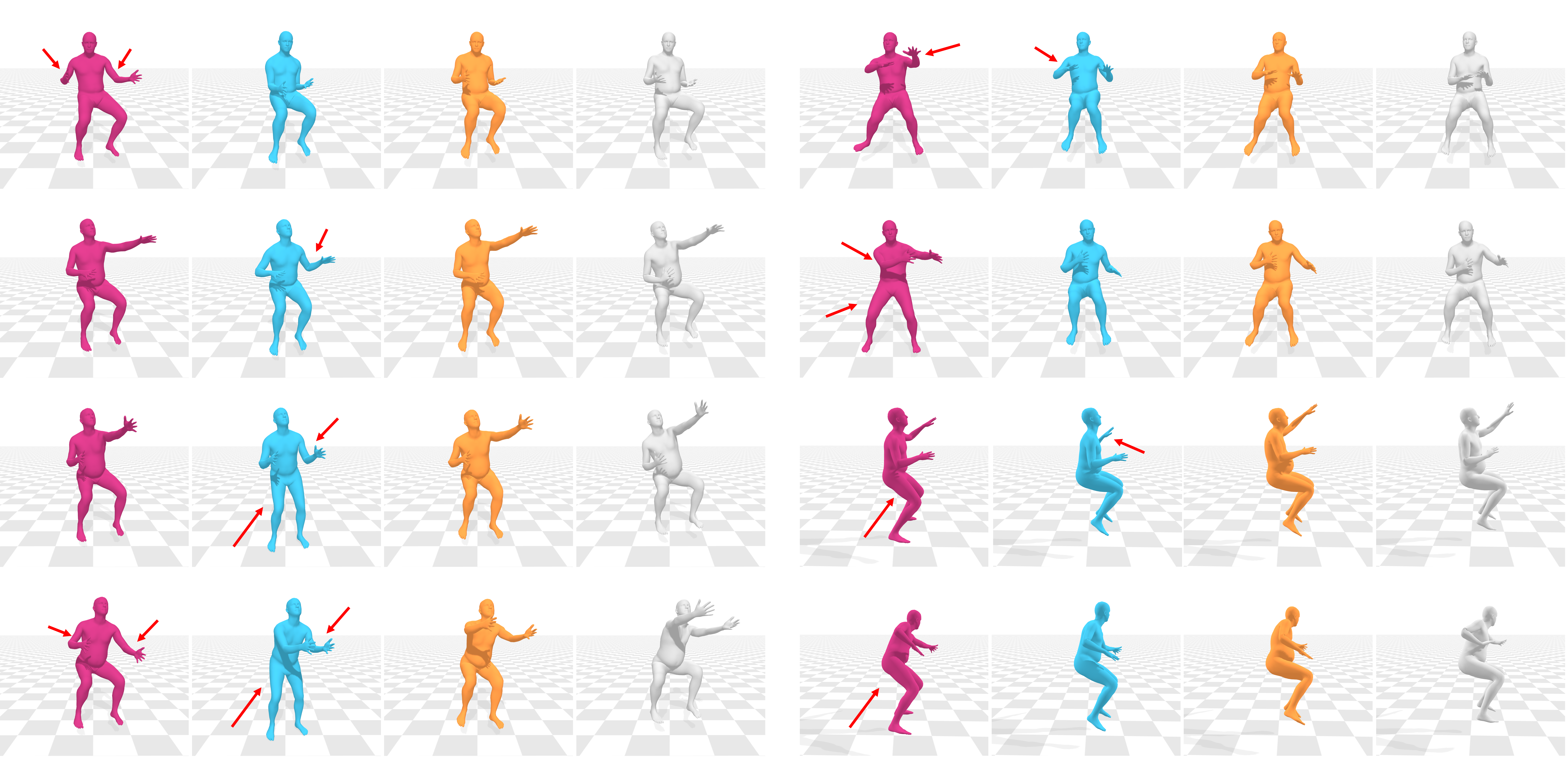}
	\caption{Additional qualitative results on DIP-IMU with previous virtual-to-real SOTAs. From left to right: TIP~\cite{jiang2022transformer}, PIP~\cite{yi2022physical}, DynaIP and GT.}
	\label{fig:dip_add}
	\vspace{-2mm}
\end{figure*}

\begin{figure*}[!t]
    \centering
	\includegraphics[width=16cm]{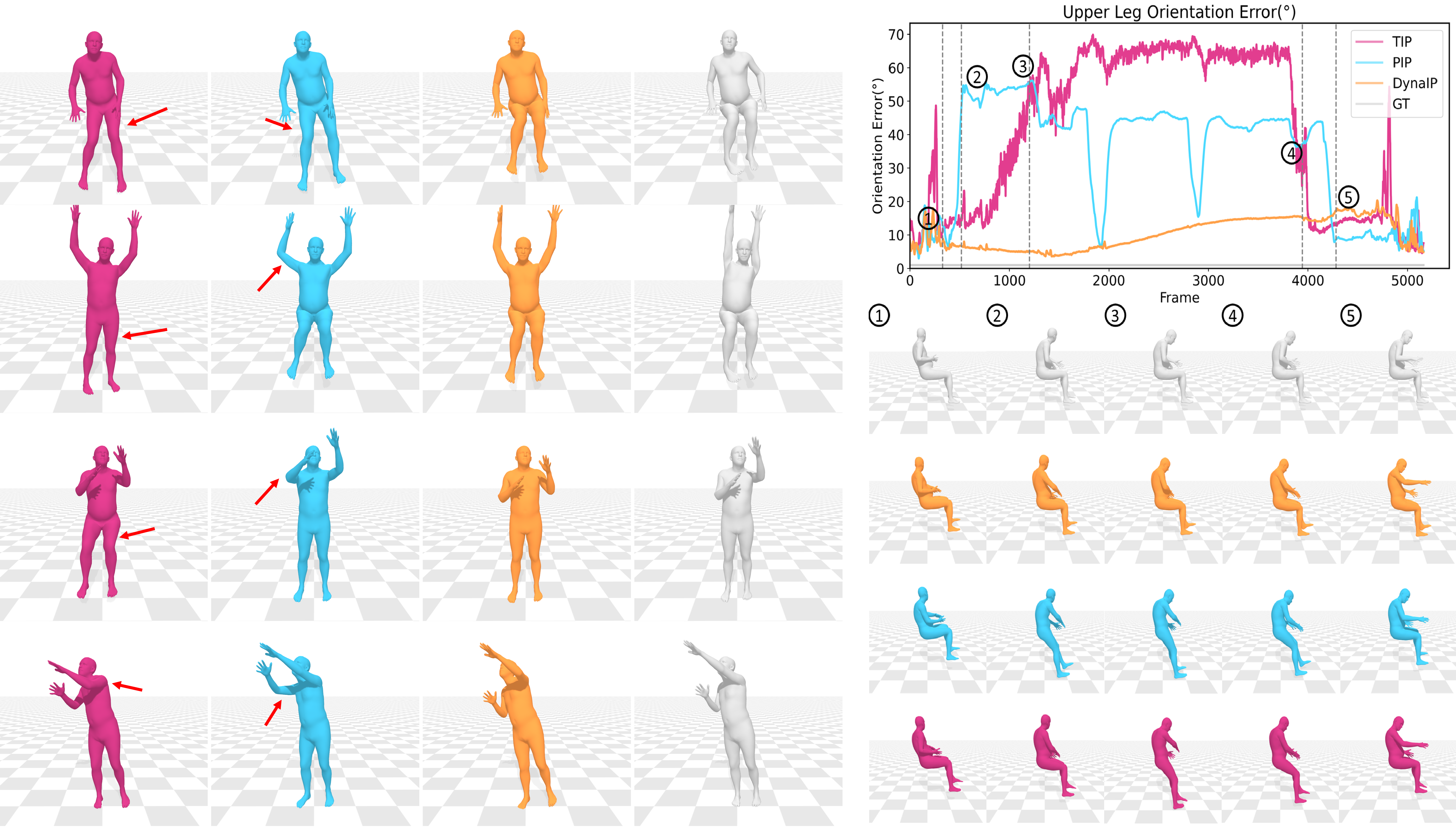}
	\caption{Additional visualization when models all trained using Xsens mocap data. Left: results from CIP sequence, Right: results from Natural Motion sequence.}
	\label{fig:xsens_add}
	\vspace{-5mm}
\end{figure*}

\section{Additional Analysis}
\label{sec:add_analysis}

In this section, we present more experimental results of our proposed method (DyanIP).


\textbf{{Additional Comparison on Unified Mocap Data and Virtual-to-Real (V-to-R) Training Scheme.}} We extend our analysis by including results from the Xsens test set for the four training settings previously discussed in Sec.~\ref{sec:experiments} on more real mocap datasets. Additionally, we report a extra setting marked as model \textit{5)}: AMASS+Xsens+DIP. This setting involves pretrain on AMASS synthetic data and then retrain with a mix of Xsens and DIP-IMU data.

The findings, as shown in Tab.~\ref{table:more_v_to_r}, reveal that relying solely on AMASS (model \textit{1)}) leads to subpar performance in real-world scenarios. Fine-tuning with a smaller real dataset, such as DIP-IMU, offers limited improvements (model \textit{2)}).

A notable observation is that the performance of models \textit{4)} and \textit{5)} is quite similar, indicating that pre-training on AMASS does not significantly affect the outcome. This similarity may stem from the test set’s focus on everyday motions, which are well-represented in the current real-world training data. However, the gap between virtual and real IMU data remains a concern and could be further investigated. Virtual data's potential may lie in capturing more diverse and uncommon motions, such as those found in sports like basketball, swimming, parkour, or specific dance styles, which are not adequately covered by existing real datasets.



\textbf{{Additional Ablation Study.}} We show the comparison results of ablation models on more datasets in Tab.~\ref{table:more_ablation_comparison}. Along with aforementioned variants (\ding{172}, \ding{173}, \ding{174}), We additionally evaluate a variant of our model denotes as "\ding{175} Complete partition (CP)", which further divides body into left and right sides: four limb groups and one torso group, as it is interesting to examine whether more local parts can bring better performance.

The results, detailed in Tab.~\ref{table:more_ablation_comparison}, demonstrate that DynaIP* maintains the best overall performance on the most test set, particularly on the DIP-IMU, Natural Motion, and CIP datasets. These datasets include more complex motion scenarios, where the unique benefits of pseudo-velocity estimation and part-based modeling are most evident.

In our experiments, the complete partition model, marked as \ding{175}, reveals a subtle decline in SIP error performance, as indicated in Tab.~\ref{table:more_ablation_comparison}. Additionally, we present the Cumulative Distribution Function (CDF) of the largest orientation error between the two upper legs on the Natural Motion test set in Fig.~\ref{fig:partition_ablation}.
While the introduction of more sensors and joint groups theoretically enhances the model's capacity to capture local motion information, over-segmentation of the body into numerous parts may inadvertently compromise the inter-connectivity between the sub-networks. This disconnect is not entirely compensated by the incorporation of low-dimensional global context. As a result, this can lead to inconsistencies in the pose estimates, manifesting as increased errors in complex motions, as depicted in Fig.~\ref{fig:partition_ablation}. Our observations suggest that while segmenting the body into local parts enhances local correlation, an excessive division may inadvertently weaken the model's ability to maintain global coherence, thus impacting its performance negatively in certain challenging scenarios.

\textbf{{Additional Comparison on Model performance.}} To isolate data impact and assess model performance, we provide a additional comparison with our model marked as DynaIP$^\dag$ using a data set consistent with PIP (combined with AMASS and DIP data) in Tab.~\ref{table:DIP_sota}. The performance enhancement highlights our architecture's efficiency.

\begin{table}[t]
\centering
\resizebox{0.43\textwidth}{!}{
\begin{tabular}{llcccc}
\toprule
\textbf{Dataset}  & \textbf{Method} & SIP Err(°) & Ang Err(°) & Pos Err(cm) & Mesh Err(cm)\\
\midrule
\multirow{2}{*}{\textbf{DIP\_IMU}} &PIP & 15.02 & 8.73 & 5.04 & \textbf{5.95}\\

& DynaIP$^\dag$ & \textbf{14.11} & \textbf{7.00} & \textbf{4.97} & 5.97\\
\midrule
\multirow{2}{*}{\begin{tabular}[c]{@{}l@{}}\textbf{Total}\\ \textbf{Capture}\end{tabular}}
&PIP & 12.93 & 12.04 &5.61 & 6.51\\

& DynaIP$^\dag$ & \textbf{12.42} & \textbf{11.06} & \textbf{5.11} & \textbf{5.79}\\
\bottomrule
\end{tabular}
}
\captionsetup{font={small}}
\caption{The performance comparison on DIP-IMU and Totalcapture. DynaIP$^\dag$ ia trained on AMASS and DIP-IMU, same as PIP.}
\label{table:DIP_sota}
\end{table}

\begin{figure}
    \centering
    \begin{subfigure}{0.15\textwidth}
        \centering
        \includegraphics[width=\textwidth]{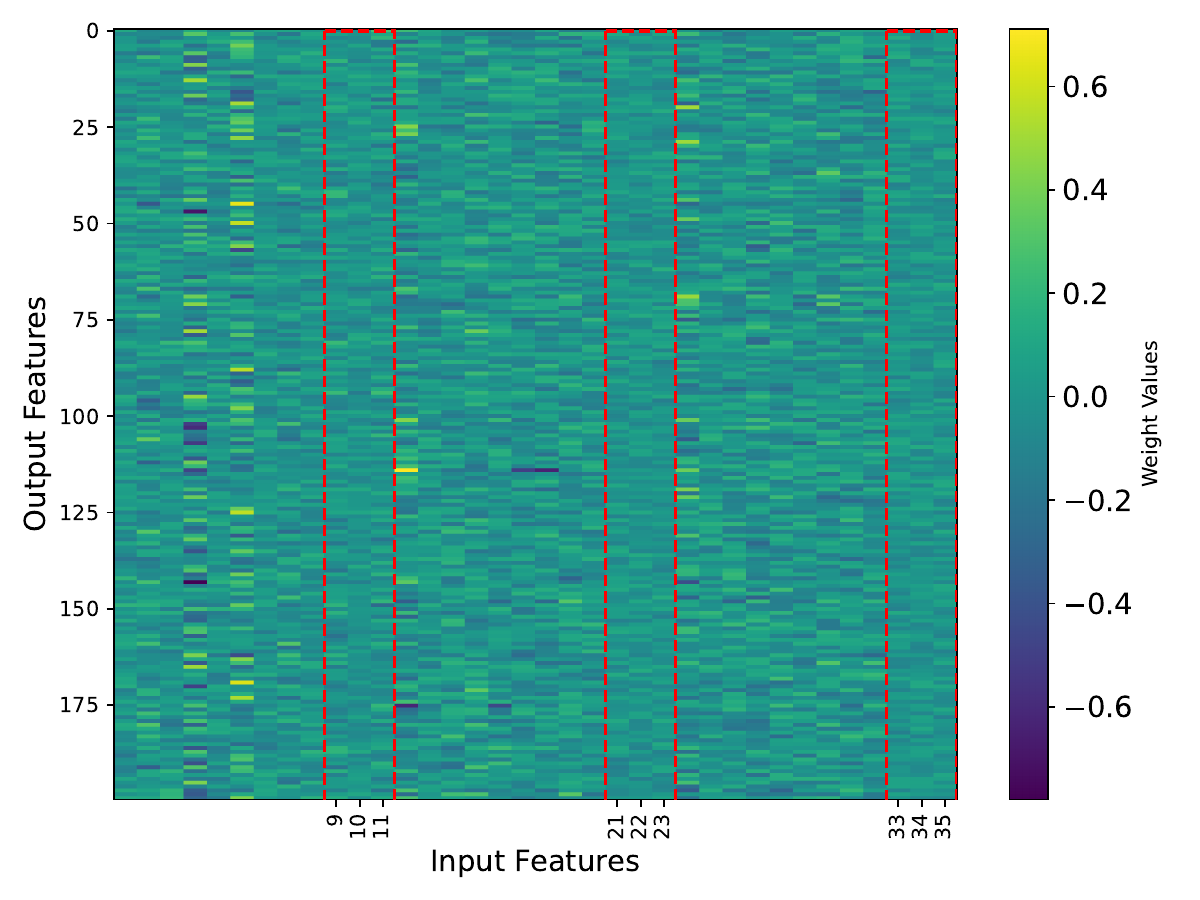}
        \caption{DynaIP*}
        \label{fig:sub1}
    \end{subfigure}
    \begin{subfigure}{0.15\textwidth}
        \centering
        \includegraphics[width=\textwidth]{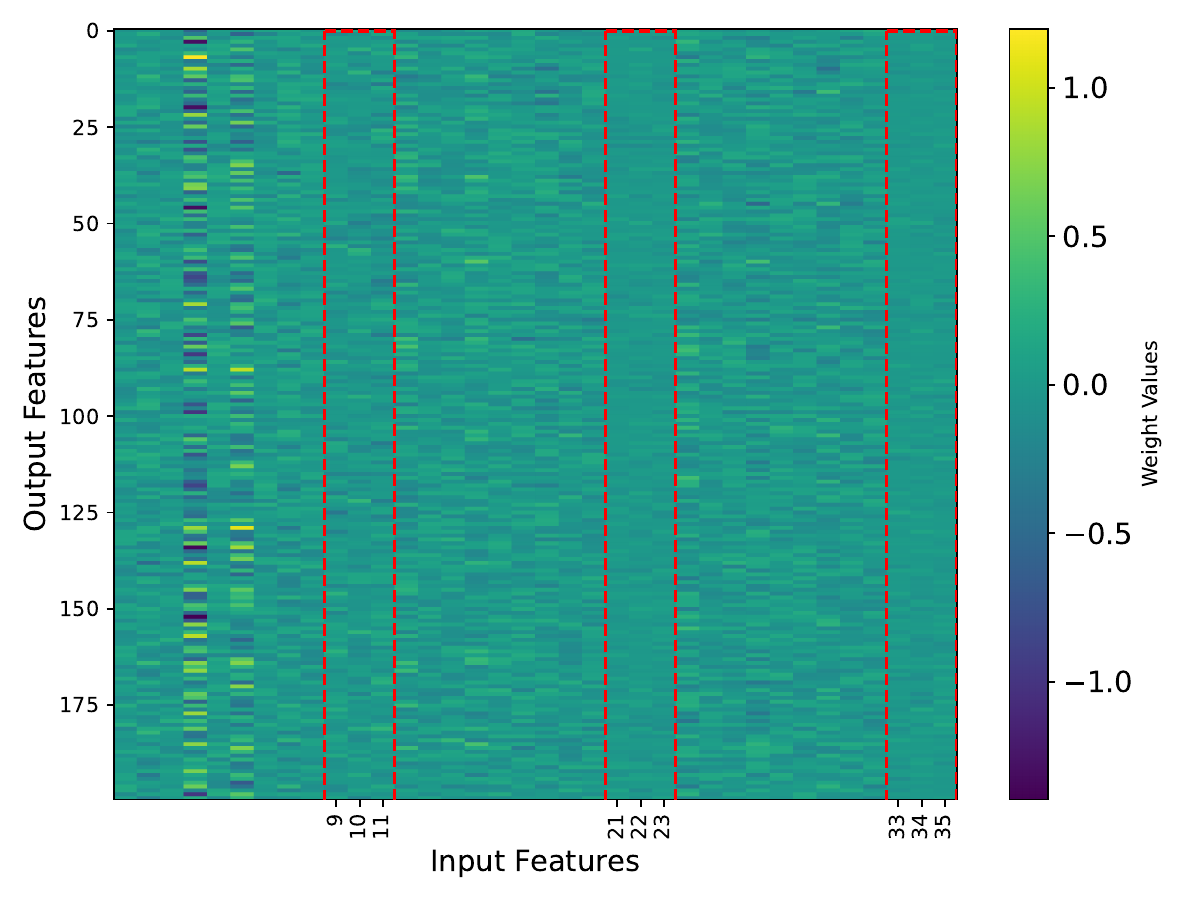}
        \caption{DynaIP*-Position}
        \label{fig:sub2}
    \end{subfigure}
    \begin{subfigure}{0.15\textwidth}
        \centering
        \includegraphics[width=\textwidth]{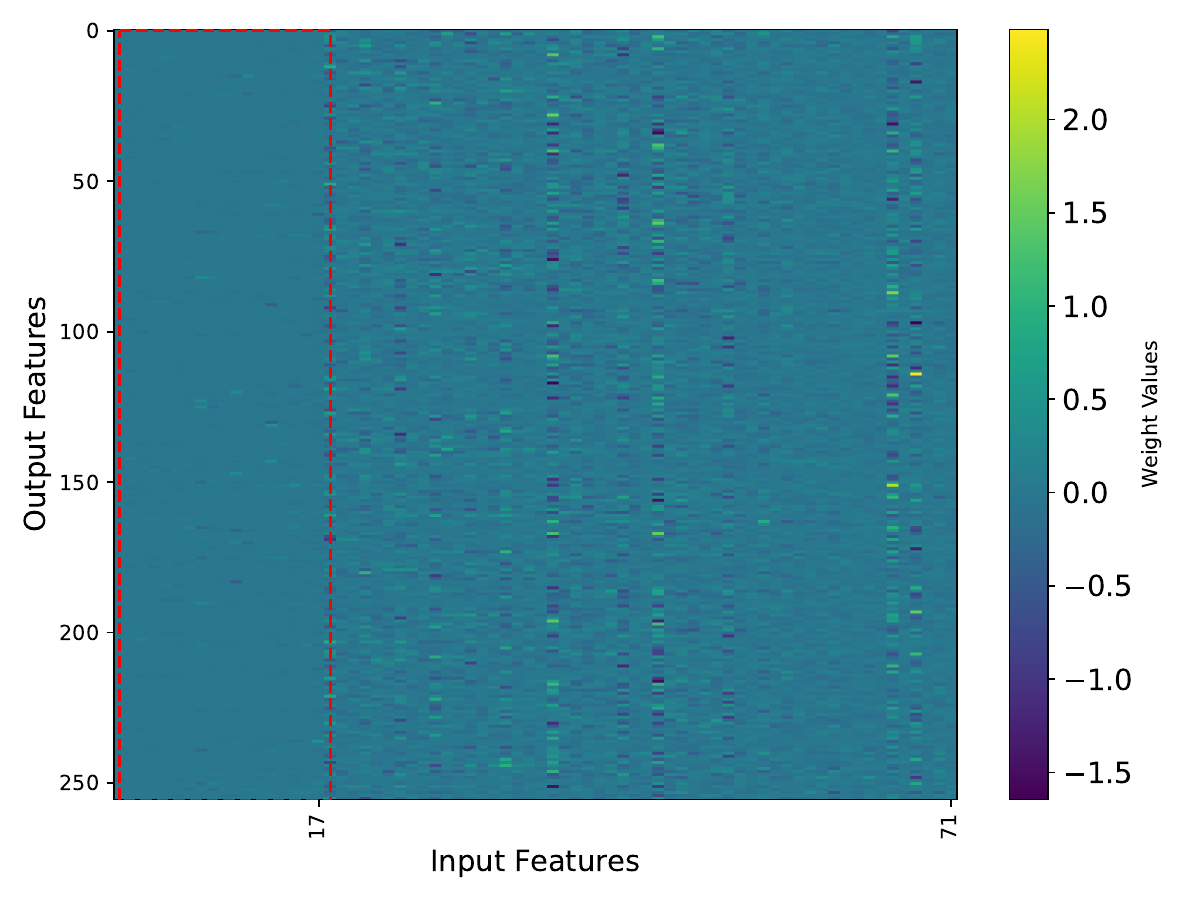}
        \caption{PIP-official}
        \label{fig:sub3}
    \end{subfigure}
    \vspace{-1mm}    \caption{Visualization of first linear layer weights, Accs on marked cols while the rest are Oris. (PIP Accs on col 0 to 17)}
    \label{fig:weights}
    \vspace{-3mm}
\end{figure}

\section{Additional Qualitative Results}
\label{sec:add_qualitative}
In this section, We show more visualization results of our method compared to the state-of-the-art methods

\textbf{Additional Qualitative Comparison with the V-to-R SOTAs.} We present a comprehensive qualitative comparison between our method, DynaIP*, and previous state-of-the-art models using virtual-to-real (V-to-R) training scheme, such as TIP~\cite{jiang2022transformer} and PIP~\cite{yi2022physical}. Additional frames from the DIP-IMU dataset are visualized in Fig.~\ref{fig:dip_add}, where our approach, DynaIP*, displays marked improvements over the previous V-to-R models in various situations, highlighting the advantages of incorporating real inertial mocap data.

This comparison points out that past V-to-R models mainly focused on real IMU datasets with SMPL ground truths. As a result, the proportion of 'real' components in these models was limited, which constrained their overall performance. By contrast, as shown in Fig.~\ref{fig:dip_add}, our approach is built on a wide array of real inertial data, which enhances the model's adaptability and precision across various motion contexts.


\textbf{Additional Qualitative Comparison on network structure.}
This visualization presents a deeper qualitative analysis that compares the performance of our model, DynaIP, with state-of-the-art models (TIP~\cite{jiang2022transformer} and PIP~\cite{yi2022physical}) trained solely on Xsens data. In Fig.~\ref{fig:xsens_add}, we exhibit more instances to highlight the enhanced capabilities of our model.

On the left side of the image, DynaIP's shows better ability to capture the movements of the upper leg and arm. This precision can be attributed to our model's advanced network structure, which effectively integrates dynamic motion information and spatial correlations within the body.

Turning to the right side of the image, our model demonstrates its robust in tracking a sitting pose, even when it involves complex hand movements. This scenario often poses a challenge for other models due to their limited capacity in localized region modeling. However, DynaIP's part-based approach allows for a more focused and accurate interpretation of motions within each region.


\textbf{Effect of using velocity as intermediate outputs.}  
Previous works, such as DIP, identified a trend where networks tend to discard much of the acceleration, and introduced an auxiliary task to reconstruct the acceleration for alleviating this problem. To better utilize acceleration, we first regress the pseudo velocity as an intermediary output, supervised by the ground-truth velocity obtained from the human pose. 
Compared with PIP/Transpose that uses leaf joints positions as intermediate outputs, our method has certain advantages in terms of using acceleration. Since noise in acceleration easily diverges with quadratic integration, it's more difficult to capture the spatial relationship in noisy acceleration. Instead, the network learns to infer the position more from the orientation input due to human kinematic relationship. Following DIP's explanation, we visualize the weights of the first linear layer of the network in the Fig.~\ref{fig:weights}, where both the weights of PIP and ours-w.position in the corresponding dimension of the acceleration input is almost zero.

\begin{figure}[!t]
    \centering
	\includegraphics[width=7cm]{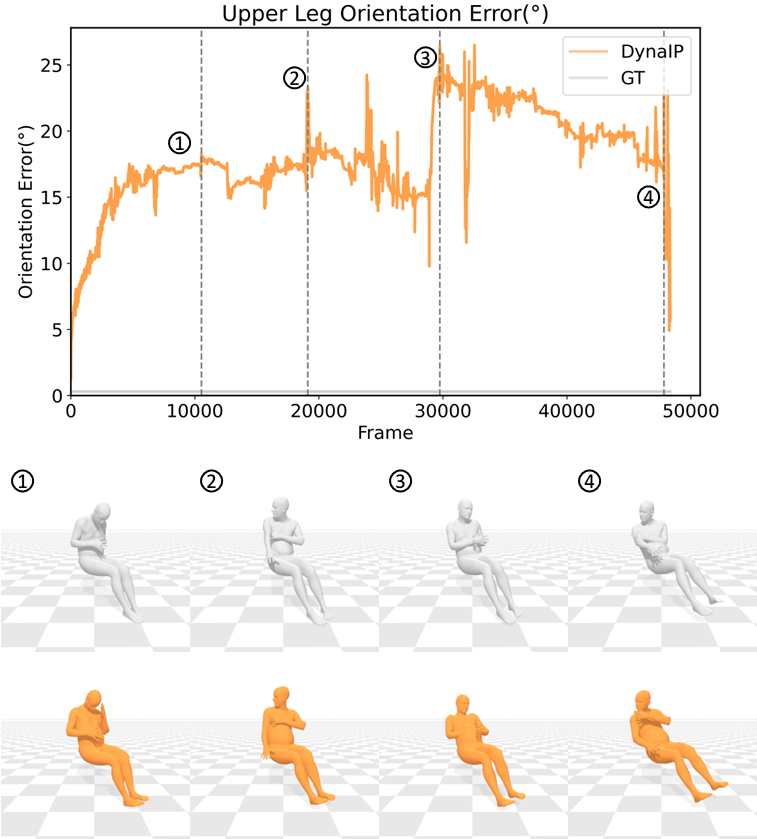}
	\caption{We've selected four timestamps with relatively high upper leg orientation errors from a 50,000 frame sitting sequence and have visualized their respective mesh representations with ground truth.}
	\label{fig:long_sitting}
	\vspace{-2mm}
\end{figure}

\textbf{Tracking robustness of long-time sitting motion.} 
In assessing the robustness of our model, DynaIP, particularly in prolonged motion scenarios, we conducted a qualitative analysis using a real sitting sequence from the Natural Motion dataset. This sequence, lasting for 14 minutes, is showcased in Fig.~\ref{fig:long_sitting} and exemplifies the model's capability in long-time motion tracking. 

The accurate modeling of a real seating posture is notably challenging. It entails not only managing the inherent noise in IMU readings but also addressing the subtle and occasional body movements humans exhibit in unscripted situations, as opposed to a perfectly stationary posture. Our model's part-based approach plays a crucial role in its success in this context.

Furthermore, to reinforce our findings and demonstrate the practical applicability of our model, we have also evaluated DynaIP in real-world scenarios using an inertial mocap device. Please refer to the supplementary videos for more details.

\section{Discussions and Future Works}
\label{sec:discussion}

To our knowledge, we are the first to utilize pre-existing publicly available inertial mocap datasets in learning-based sparse inertial mocap with the global orientation mapping strategy across skeleton formats. Despite our efforts to expand the training data with real IMU readings, we acknowledge a challenge: the model's accuracy dips when it encounters poses that are underrepresented in the training set. To address this, we recognize the potential of integrating additional datasets featuring real IMUs, which cover a broader spectrum of motion types. Recently, there comes several multi-modality datasets such as~\cite{dai2023sloper4d, dai2022hsc4d} that make extensive use of IMU data as part of their sensor modality. Regrettably, raw IMU data synchronized with these resources aren't presently available for public access. We are of the strong belief that, once made available, these datasets could offer immense value for inertial sensor-based motion capture. 

At present, our system does not incorporate improvements in global root translation and operates based on the assumption of flat ground conditions. This limitation stems from the inherent tendency of purely inertial-based global trajectory estimation methods to experience drift. Additionally, the accuracy of our current methods for estimating translation can be adversely impacted by inaccuracies in pose prediction, a problem also linked to the susceptibility of inertial-based systems to drift. Looking ahead, we aim to tackle the issue of long-term drift in IMU-based solutions by integrating environmental constraint information or by incorporating additional sensor modalities.


In summary, our research points towards exciting avenues for future work in sparse IMU-based human pose estimation. Expanding the dataset diversity, dividing distinct body regions, and enhancing velocity estimation are key areas that hold the potential to advance this field significantly. As we continue to explore these possibilities, we would also be committed to using widely used smart devices containing IMUs, such as VR \cite{jiang2022avatarposer, yang2021lobstr}, mobile phones, smart-watches \cite{mollyn2023imuposer}, etc., to develop robust and accurate motion capture solutions that can thrive in a variety of real-world applications.

\end{document}